\documentclass[preprint,12pt]{elsarticle}

\usepackage{amssymb}
\usepackage{amsmath}
\usepackage{xcolor}
\definecolor{red3}{HTML}{C52A20}
\definecolor{bluex}{HTML}{6f94e7}

\usepackage{hyperref}
\hypersetup{hidelinks}
\usepackage{algorithm,algpseudocode,subfig}

\newcommand{\vtheta}{{\boldsymbol \theta}}
\newcommand{\vpsi}{{\boldsymbol \psi}}
\newcommand{\hD}{\mathcal{D}}
\newcommand{\hB}{\mathcal{B}}
\newcommand{\vg}{{\bf g}}
\newcommand{\bE}{\mathbb{E}}
\newcommand{\hA}{\mathcal{A}}

\usepackage[inline, shortlabels]{enumitem}

\usepackage{graphicx, multirow, booktabs, bm, xcolor, bbding, physics}

\usepackage{amsthm}
\newtheorem{theorem}{Theorem}[section]
\newtheorem{proposition}[theorem]{Proposition}

\theoremstyle{definition}

\theoremstyle{remark}

\definecolor{teal}{HTML}{508AB2}

\journal{Neural Networks}

\begin{document}
	
	\begin{frontmatter}
		
		%% Title, authors and addresses
		
		%% use the tnoteref command within \title for footnotes;
		%% use the tnotetext command for theassociated footnote;
		%% use the fnref command within \author or \affiliation for footnotes;
		%% use the fntext command for theassociated footnote;
		%% use the corref command within \author for corresponding author footnotes;
		%% use the cortext command for theassociated footnote;
		%% use the ead command for the email address,
		%% and the form \ead[url] for the home page:
		%% \title{Title\tnoteref{label1}}
		%% \tnotetext[label1]{}
		%% \author{Name\corref{cor1}\fnref{label2}}
		%% \ead{email address}
		%% \ead[url]{home page}
		%% \fntext[label2]{}
		%% \cortext[cor1]{}
		%% \affiliation{organization={},
			%%             addressline={},
			%%             city={},
			%%             postcode={},
			%%             state={},
			%%             country={}}
		%% \fntext[label3]{}
		
		\title{Dual-Balancing for Multi-Task Learning} %% Article title
		
		%% use optional labels to link authors explicitly to addresses:
		%% \author[label1,label2]{}
		%% \affiliation[label1]{organization={},
			%%             addressline={},
			%%             city={},
			%%             postcode={},
			%%             state={},
			%%             country={}}
		%%
		%% \affiliation[label2]{organization={},
			%%             addressline={},
			%%             city={},
			%%             postcode={},
			%%             state={},
			%%             country={}}
		
		\author[label1,label2]{Baijiong Lin}\ead{bj.lin.email@gmail.com}
		\author[label3]{Weisen Jiang}
		\author[label4]{Feiyang Ye}
		\author[label4]{Yu Zhang}
		\author[label5]{Pengguang Chen}
		\author[label1,label2,label6]{Ying-Cong Chen\corref{cor1}}\ead{yingcongchen@ust.hk}
		\author[label5]{Shu Liu\corref{cor1}}\ead{liushuhust@gmail.com}
		\author[label7]{Ivor W. Tsang}
		\author[label6]{James T. Kwok}
		
		%% Author affiliation
		\affiliation[label1]{organization={The Hong Kong University of Science and Technology (Guangzhou)},
			city={Guangzhou},
			postcode={510000},
			country={China}
		}
		
		\affiliation[label2]{organization={HKUST(GZ) - SmartMore Joint Lab},
			city={Guangzhou},
			postcode={510000},
			country={China}
		}
		
		\affiliation[label3]{organization={The Chinese University of Hong Kong},
			city={Hong Kong},
			postcode={999077},
			country={China}
		}
		
		\affiliation[label4]{organization={Southern University of Science and Technology},
			city={Shenzhen},
			postcode={518055},
			country={China}
		}
		
		\affiliation[label5]{organization={SmartMore},
			city={Shenzhen},
			postcode={518000},
			country={China}
		}
		
		\affiliation[label6]{organization={The Hong Kong University of Science and Technology},
			city={Hong Kong},
			postcode={999077},
			country={China}
		}
		
		\affiliation[label7]{organization={Centre for Frontier AI Research, A$^*$STAR},
			postcode={138632},
			country={Singapore}
		}
		
		\cortext[cor1]{Corresponding authors.}
		
		%% Abstract
		\begin{abstract}
			Multi-task learning aims to learn multiple related tasks simultaneously and has achieved great success in various fields. However, the disparity in loss and gradient scales among tasks often leads to performance compromises, and the balancing of tasks remains a significant challenge. In this paper, we propose Dual-Balancing Multi-Task Learning (DB-MTL) to achieve task balancing from both the loss and gradient perspectives. Specifically, {DB-MTL achieves loss-scale balancing by performing logarithm transformation on each task loss, and rescales gradient magnitudes by normalizing all task gradients to comparable magnitudes using the maximum gradient norm.} Extensive experiments on a number of benchmark datasets demonstrate that DB-MTL consistently performs better than the current state-of-the-art.
		\end{abstract}
		
		%%%Graphical abstract
		%\begin{graphicalabstract}
		%%\includegraphics{grabs}
		%\end{graphicalabstract}
		%
		%%%Research highlights
		%\begin{highlights}
		%\item Research highlight 1
		%\item Research highlight 2
		%\end{highlights}
		
		%% Keywords
		
	\end{frontmatter}
	
	\section{Introduction}
	Multi-task learning (MTL) \cite{caruana1997multitask, zhang2021survey, chen2025modl} jointly learns multiple related tasks using a single model, improving parameter-efficiency
	and inference speed compared to learning a separate model for each task. By sharing the model, MTL can extract common knowledge to improve each task's performance. It has demonstrated its superiority in various fields, such as computer vision \cite{vandenhende2021multi, ye2022inverted,lin2024mtmamba,lin2024mtmambaplus,luo2025hirmtl}, natural language processing \cite{liu2017adversarial, liu2019multi, sun2020learning, chen2024multi, wang2021gradient}, and recommendation systems \cite{tang2020progressive, hazimeh2021dselect, wang2023multi,yi2025hybrid}.
	
	To learn multiple tasks simultaneously, equal weighting (EW) \cite{zhang2021survey} is a straightforward method that minimizes the sum of task losses with equal task weights. However, it usually suffers from the challenging
	\textit{task balancing problem} \cite{vandenhende2021multi, linreasonable}, in which some tasks perform well while others do not \cite{standley2020tasks}. To alleviate this problem, a number of methods have been recently proposed by dynamically tuning the task weights. They can be categorized as \textit{loss balancing} \cite{kendall2018multi, ljd19, ye2021multi,ye2024moml,
		ye2024first, liuauto} and \textit{gradient balancing} \cite{chen2018gradnorm, sk18, chen2020just, yu2020gradient, liu2021conflict, liu2021imtl, wang2021gradient, navon2022multi,
		fernando2023mitigating}. Loss balancing methods balance the tasks based on the learning speed \cite{ljd19} or validation performance \cite{ye2021multi, ye2024first, liuauto} \textit{at the loss level}, while gradient balancing methods balance the gradients by mitigating gradient conflicts \cite{yu2020gradient} or enforcing gradient norms to be close \cite{chen2018gradnorm} \textit{at the gradient level}. However, recently, multiple
	extensive empirical studies \cite{linreasonable,kurindefense,xincurrent} demonstrate that the performance of these existing methods is still unsatisfactory,
	indicating that task balancing is still an open problem.
	
	To mitigate the task balancing problem, in this paper, we consider simultaneously balancing both the loss scales (at the loss level) and gradient magnitudes (at the gradient level).
	Since the loss scales/gradient magnitudes among tasks can be different, those with large values can dominate the update direction of the model, causing unsatisfactory performance on some other tasks \cite{standley2020tasks, liu2021imtl}. Therefore, we propose a simple yet effective Dual-Balancing Multi-Task Learning (\textbf{DB-MTL}) method that consists of both loss-scale and gradient-magnitude balancing. First, we perform a logarithm transformation on each task loss to make all task losses have a similar scale. This is non-parametric and can recover the loss transformation in IMTL-L \cite{liu2021imtl}. We find that the logarithm transformation also benefits existing gradient balancing methods. Second, we normalize all task gradients to the same magnitude as the maximum gradient norm. This is training-free and guarantees all gradients' magnitude are the same compared with GradNorm \cite{chen2018gradnorm}.  Empirically, we find that the magnitude of normalized gradients plays an important role in performance, and setting it as the maximum gradient norm among tasks performs the best. Extensive experiments are performed on a number of benchmark datasets. Results demonstrate that DB-MTL consistently outperforms the current state-of-the-art. 
	
	Our contributions can be summarized as follows:
	\begin{enumerate}
		\item We propose DB-MTL, a novel dual-balancing approach that simultaneously addresses both loss-scale and gradient-magnitude imbalances in multi-task learning through:
		\begin{itemize}
			\item A parameter-free logarithm transformation for loss-scale balancing that effectively equalizes loss scales across tasks;
			\item {A maximum-norm gradient normalization strategy that rescales all task gradients to comparable magnitudes for balanced model updates.}
		\end{itemize}
		\item We conduct extensive experiments across diverse benchmarks demonstrating that DB-MTL consistently outperforms state-of-the-art MTL methods.
	\end{enumerate}
	
	{\paragraph{Notations} For clarity, we summarize the key notations used throughout this paper. We use $T$ to denote the number of tasks, $\mathcal{D}_t$ for the training dataset of task $t$, $\vtheta$ and $\{\vpsi_t\}_{t=1}^T$ for task-sharing and task-specific parameters respectively, $\gamma_t$ for task weights, and $\ell_t$ for the loss function of task $t$. $\vg_{t,k}$ and $\tilde{\vg}_k$ represent the gradient and aggregated gradient at iteration $k$, with $\alpha_k$ as the scaling factor.}
	
	\section{Related Works}
	In an MTL problem with $T$ tasks, we aim to learn a model from $\{\mathcal{D}_t\}_{t=1}^T$, where $\mathcal{D}_t$ is the training dataset of task $t$. The MTL model parameters can be divided into two parts: (i) task-sharing parameter $\vtheta$, and (ii) task-specific parameters $\{\vpsi_t\}_{t=1}^{T}$. For example, in computer vision tasks, $\vtheta$ usually represents a feature
	encoder (e.g., \textit{ResNet} \cite{he2016deep}) to extract common features among tasks, while $\vpsi_t$ corresponds to the task-specific output module (e.g., a fully-connected
	layer). For parameter efficiency, $\vtheta$ contains most of the MTL model parameters, and is crucial to the performance.
	
	Let $\ell_t(\mathcal{D}_t;\vtheta,\vpsi_t)$ be the loss on 
	task $t$'s data $\mathcal{D}_t$ using parameter $(\vtheta, \vpsi_t)$. The training objective of MTL is $\sum_{t=1}^T \gamma_t \ell_t(\mathcal{D}_t; \vtheta, \vpsi_t)$, where $\gamma_t$ is the weight for task $t$. Equal weighting (EW) \cite{zhang2021survey} is a simple MTL approach that sets $\gamma_t=1$ for all tasks. However, EW usually suffers from the task balancing problem in which some tasks have unsatisfactory
	performance \cite{standley2020tasks}. To improve its performance,
	many other MTL methods have been proposed to dynamically tune the task weights $\{\gamma_t\}_{i=1}^T$ during training. They can be categorized as loss balancing, gradient balancing, or hybrid balancing.
	
	\vspace{-0.04in}
	\subsection{Loss Balancing Methods} 
	This approach weights the task losses with $\{\gamma_t\}_{i=1}^T$ that are computed dynamically. $\{\gamma_t\}_{i=1}^T$ affect the update of both the task-sharing parameter $\vtheta$ and task-specific parameter $\{\vpsi_t\}_{t=1}^{T}$. They can be set based on measures such as homoscedastic uncertainty \cite{kendall2018multi}, learning speed \cite{ljd19}, validation performance \cite{ye2021multi, ye2024first}, and improvable gap \cite{dai2023improvable}. Alternatively, IMTL-L \cite{liu2021imtl} encourages the weighted losses $\{\gamma_t\ell_t(\mathcal{D}_t; \vtheta, \vpsi_t)\}_{t=1}^T$ to have similar loss scale across all tasks by transforming each loss $\ell_t(\mathcal{D}_{t};\vtheta, \vpsi_{t})$ as $e^{s_t}\ell_t(\mathcal{D}_{t};\vtheta, \vpsi_{t})-s_t$, where $\{s_t\}_{t=1}^T$ are learnable parameters and obtained by gradient descent at each iteration. 
	
	\vspace{-0.04in}
	\subsection{Gradient Balancing Methods} 
	The update of the task-sharing parameter $\vtheta$ depends on all task gradients $\{\nabla_{\vtheta}\ell_t(\mathcal{D}_{t};\vtheta, \vpsi_{t})\}_{t=1}^T$. Thus, gradient balancing methods aim to aggregate all task gradients in different manners.
	For example, MGDA \cite{sk18} formulates MTL as a multi-objective optimization problem and selects the aggregated gradient with the minimum norm \cite{desideri12}.
	CAGrad \cite{liu2021conflict} improves MGDA by constraining the aggregated gradient to be around the average gradient. 
	MoCo \cite{fernando2023mitigating} mitigates the bias in MGDA by introducing a momentum-like gradient estimate and a regularization term.
	GradNorm \cite{chen2018gradnorm} learns task weights to scale the task gradients to similar magnitudes.
	PCGrad \cite{yu2020gradient} projects the gradient of one task onto the normal plane of the other if their gradients conflict.
	GradVac \cite{wang2021gradient} aligns the gradients regardless of whether the gradients conflict or not.
	GradDrop \cite{chen2020just} randomly masks out gradient values with inconsistent signs.
	IMTL-G \cite{liu2021imtl} learns task weights to enforce the aggregated gradient to have equal projections on each task gradient.
	Nash-MTL \cite{navon2022multi} formulates gradient aggregation as a Nash bargaining game. 
	
	For most gradient balancing methods (such as PCGrad \cite{yu2020gradient}, CAGrad \cite{liu2021conflict}, MoCo \cite{fernando2023mitigating}, GradDrop \cite{chen2020just}, and IMTL-G \cite{liu2021imtl}), the task weight $\gamma_t$ only affects update of the task-sharing parameter $\vtheta$, while in some other gradient balancing methods (such as MGDA \cite{sk18}, GradNorm \cite{chen2018gradnorm}, and Nash-MTL \cite{navon2022multi}), the task weight $\gamma_t$ affects the update of both the task-sharing and task-specific parameters.
	
	\subsection{Hybrid Balancing Methods} \label{sec:hybrid}
	As loss balancing and gradient balancing are complementary, these two types of methods can be combined to achieve better performance. In this approach, the task weight $\gamma_t$ is obtained as the product of the loss and gradient balancing
	weights. For example, the first hybrid balancing method
	IMTL \cite{liu2021imtl} combines IMTL-L with IMTL-G.
	Subsequently, various combinations  \cite{linreasonable,dai2023improvable,liuauto} of loss/gradient balancing methods demonstrate performance improvements. In this paper, we propose DB-MTL that combines the logarithm transformation (for loss balancing) and the maximum-norm gradient normalization (for gradient balancing).
	
	\section{Proposed Method}
	
	In this section, we alleviate the task balancing problem from both the loss and gradient perspectives. First, we balance all loss scales by performing logarithm transformation on each task's loss (Section \ref{sec:loss}). 
	Next, we achieve gradient-magnitude balancing by normalizing each task's gradient to the same magnitude as the maximum gradient norm (Section \ref{sec:si-g}).
	The procedure, called DB-MTL (Dual-Balancing Multi-Task Learning), is shown in Algorithm \ref{alg:method}. 
	
	\subsection{Scale-Balancing Loss Transformation}
	\label{sec:loss}
	
	Tasks with different types of loss functions usually have different scales, leading to the task balancing problem. For example, in the \textit{NYUv2} dataset \cite{silberman2012indoor}, the cross-entropy loss, $L_1$ loss, and cosine loss are used as the loss functions of the semantic segmentation, depth estimation, and surface normal prediction tasks, respectively. As observed in \cite{standley2020tasks, yu2020gradient, navon2022multi} and also in our experimental results in Tables \ref{tbl:mtl-nyu_dmtl} and \ref{tbl:mtl-nyu_segnet}, surface normal prediction is affected by the other two tasks (semantic segmentation and depth estimation), causing MTL methods like EW to perform unsatisfactorily.
	
	When prior knowledge of the loss scales is available, we can choose $\{s_t^\star\}_{t=1}^T$ such that $\{s_t^\star\ell_t(\hD_t; \vtheta, \vpsi_t)\}_{t=1}^T$ have the same scale, and then minimize the total loss $\sum_{t=1}^{T}s_t^\star \ell_t(\hD_t; \vtheta, \vpsi_t)$.
	Previous methods \cite{kendall2018multi, ljd19, liu2021imtl, ye2021multi} implicitly learn $\{s_t^\star\}_{t=1}^T$ when learning the task weights $\{\gamma_t\}_{t=1}^T$.
	However, obviously the optimal $\{s_t^\star\}_{t=1}^T$ cannot be obtained during training.
	
	Without the availability of $\{s_t^\star\}_{t=1}^T$, the logarithm transformation can be used to alleviate the loss scale problem. Specifically, we transform each task's loss $\ell_t(\hD_t;\vtheta, \vpsi_t)$ to $\log \ell_t(\hD_t;\vtheta, \vpsi_t)$, and then minimize $\sum_{t=1}^{T}\log \ell_t(\hD_t;\vtheta, \vpsi_t)$. Since $\log (\cdot)$ can compress the range of its input, it can reduce the loss scale gap between different tasks.
	
	IMTL-L \cite{liu2021imtl} tackles the loss scale issue using a transformed loss $e^{s_t}\ell_t(\mathcal{D}_{t};\vtheta, \vpsi_{t})-s_t$, where $s_t$ is a learnable parameter for the $t$-th task and approximately solved by one-step gradient descent at every iteration. The following Proposition \ref{pp:log} shows that IMTL-L is equivalent to the logarithm transformation when $s_t$ is the exact minimizer in each iteration. 
	
	\begin{proposition}
		For $x>0$,  $\log(x) = \min_s e^s x - s - 1$.
		\label{pp:log}
	\end{proposition}
	\noindent
	\begin{proof}
		Define an auxiliary function $f(s) = e^s x - s - 1$. 
		It is easy to show that $\dv{f(s)}{s}=e^s x -1$ and $\dv[2]{f(s)}{s}=e^s x>0$. Thus, $f(s)$ is convex. By the first-order optimal condition~\cite{Boyd2004}, let $e^{s^\star} x -1=0$, the global minimizer is solved as $s^\star = -\log(x)$. Therefore, $f(s^\star)=e^{s^\star} x - s^\star - 1=e^{-\log(x)}x+\log(x)-1=\log(x)$, where we finish the proof.
	\end{proof}
	
	Compared to IMTL-L, the logarithm transformation does not require additional parameters and computational cost during training. Thus, the logarithm transformation is simpler and more effective than IMTL-L.
	
	\begin{algorithm}[!t]
		\caption{Dual-Balancing Multi-Task Learning.}
		\label{alg:method}
		\begin{algorithmic}[1]
			\Require numbers of iterations $K$, learning rate $\eta$, tasks $\{\mathcal{D}_t\}_{t=1}^T$, $\epsilon=10^{-8}$, $\beta$; 
			\State randomly initialize $\vtheta_0, \{\vpsi_{t, 0}\}_{t=1}^T$; 
			\State initialize $\hat{\vg}_{t,-1}=\bf{0}$, for all $t$;
			\For{$k=0,\dots, K-1$}
			\For{$t=1,\dots, T$}
			\State sample a mini-batch dataset $\mathcal{B}_{t,k}$ from $\mathcal{D}_t$; \label{alg-step: sample}
			\State $\vg_{t,k}=\nabla_{\vtheta_{k}}\log(\ell_t({\mathcal{B}}_{t,k};\vtheta_k, \vpsi_{t,k})+\epsilon)$;
			\label{alg-step: task grad}
			\State compute $\hat{\vg}_{t,k} = \beta  \hat{\vg}_{t,k-1} + (1 - \beta)  \vg_{t,k}$; 
			\label{alg-step: ema}
			\EndFor
			\State compute $\tilde{\vg}_k = \alpha_k \sum_{t=1}^T \frac{\hat{\vg}_{t,k}}{\|\hat{\vg}_{t,k}\|_2+\epsilon}  $, where $\alpha_k = \max_{1\leq t \leq T} \|\hat{\vg}_{t, k}\|_2$;  \label{alg-step: compute g-k}
			\State update task-sharing parameter by $\vtheta_{k+1}=\vtheta_k-\eta \tilde{\vg}_k$; \label{alg-step: share-update}
			\For{$t=1,\dots, T$} \label{alg-step:task-specifc update-a}
			\State $\vpsi_{t,k+1}=\vpsi_{t,k}-\eta\nabla_{\vpsi_{t,k}}\log(\ell_t({\mathcal{B}}_{t,k};\vtheta_k, \vpsi_{t,k})+\epsilon)$;
			\EndFor \label{alg-step:task-specifc update-b}
			\EndFor
			\State \textbf{Return} $\vtheta_{K}, \{\vpsi_{t, K}\}_{t=1}^T$.
		\end{algorithmic}
	\end{algorithm}
	
	\subsection{Magnitude-Balancing Gradient Normalization} \label{sec:si-g}
	In addition to the task losses, task gradients also suffer from the scale issue. As the update direction of $\vtheta$ is obtained by uniformly averaging all task gradients, it may be dominated by the large task gradients, causing sub-optimal performance \cite{yu2020gradient, liu2021conflict}.
	
	A simple approach is to normalize task gradients to the same magnitude. As computing the batch gradient is computationally expensive, mini-batch stochastic gradient descent is often used in practice. Specifically, at iteration $k$, we sample a mini-batch $\hB_{t,k}$ from $\hD_t$ for the $t$-th task (step \ref{alg-step: sample} in Algorithm \ref{alg:method}) and compute the mini-batch gradient ${\vg}_{t,k} = \nabla_{\vtheta_{k}}\log\ell_t(\mathcal{B}_{t,k};\vtheta_{k}, \vpsi_{t,k})$ (step \ref{alg-step: task grad} in Algorithm \ref{alg:method}). 
	Exponential moving average (EMA), which is popularly used in adaptive gradient methods (e.g., RMSProp~\cite{tieleman2012lecture}, AdaDelta~\cite{zeiler2012adadelta}, and Adam~\cite{kingma2015adam}), is used to estimate $\bE_{\hB_{t,k}\sim \hD_t}\nabla_{\vtheta_{k}} \log\ell_t(\hB_{t,k}; \vtheta_{k}, \vpsi_{t,k})$ dynamically (step \ref{alg-step: ema} in Algorithm \ref{alg:method}) as
	\begin{align}
		\hat{\vg}_{t,k} = \beta \hat{\vg}_{t,k-1} + (1 - \beta) {\vg}_{t,k}, \label{eq:beta}
	\end{align}
	where $\beta\in(0,1)$ controls the forgetting rate. After obtaining the task gradients $\{\hat{\vg}_{t,k}\}_{t=1}^T$, we normalize them to have the same $\ell_2$-norm, and compute the aggregated gradient as
	\begin{align}
		\tilde{\vg}_k = \alpha_k \sum_{t=1}^T \frac{\hat{\vg}_{t,k}}{\|\hat{\vg}_{t,k} \|_2},
		\label{eq:si-g}
	\end{align}
	where $\alpha_k$ is a scaling factor controlling the update magnitude. {After normalization, all tasks contribute with comparable magnitudes to the update direction.} 
	
	The choice of $\alpha_k$ is critical in alleviating the task balancing problem. Intuitively, when some tasks have large gradient norms and others have small gradient norms, the first group of tasks has not yet converged while the second group of tasks has almost converged. The current model $\vtheta_{k}$ is undesirable and can cause the task balancing problem as not all tasks have converged. Hence, $\alpha_k$ should be large to escape this undesirable solution. On the other hand, when all task gradient norms are small, model $\vtheta_{k}$ is close to a stationary solution for all tasks, and $\alpha_k$ should be small so that the solution will no longer change.
	Thus, we choose $\alpha_k = \max_{1\leq t \leq T}\|\hat{\vg}_{t, k}\|_2$, i.e., $\alpha_k$ is small if and only if all the task gradient norms are small.  
	
	After scaling the losses and gradients, the task-sharing parameter is updated as $\vtheta_{k+1}=\vtheta_k-\eta \tilde{\vg}_k$ (step \ref{alg-step: share-update}), where  $\eta>0$ is the learning rate. For the task-specific parameters
	$\{\vpsi_{t,k}\}_{t=1}^T$, as the update of each of them only depends on the corresponding task gradient separately, their gradients do not suffer from the gradient scaling issue. Hence, the update for task-specific parameters is simply
	$\vpsi_{t,k+1}=\vpsi_{t,k}-\eta\nabla_{\vpsi_{t,k}}\log\ell_t({\mathcal{B}}_{t,k};\vtheta_k, \vpsi_{t,k})$ (steps \ref{alg-step:task-specifc update-a}-\ref{alg-step:task-specifc update-b}).
	
	GradNorm \cite{chen2018gradnorm} also aims to learn $\{\gamma_t\}_{t=1}^T$ so that the scaled gradients have similar norms. However, it has two problems. First, alternating the updates of model parameters and task weights cannot guarantee all task gradients have the same magnitude in each iteration. Second, as will be seen from Figure \ref{fig:alpha_k} in Section
	\ref{sec:sensitivity_analysis}, the choice of the update magnitude $\alpha_k$ can significantly affect performance. However, this is not considered in GradNorm.  
	
	\section{Experiments}
	\label{sec:expt}
	
	In this section, we empirically evaluate the proposed DB-MTL on a number of tasks, including scene understanding (Section~\ref{sec:nyu}), molecular property prediction (Section~\ref{sec:mo}), and image classification (Section~\ref{sec:image}).
	
	\subsection{Evaluation on Scene Understanding}
	\label{sec:nyu}
	
	\paragraph{Datasets} 
	Following RLW \cite{linreasonable}, CAGrad \cite{liu2021conflict}, and Nash-MTL \cite{navon2022multi}, the following two scene understanding datasets are used: 
	\begin{enumerate*}[(i), series = tobecont, itemjoin = \quad]
		\item \textit{NYUv2} \cite{silberman2012indoor}, which is an indoor scene understanding dataset. It has $3$ tasks ($13$-class semantic segmentation, depth estimation, and surface normal prediction) with $795$ training and $654$ testing images.
		\item \textit{Cityscapes} \cite{CordtsORREBFRS16}, which
		is an urban scene understanding dataset. It has $2$ tasks ($7$-class semantic segmentation and depth estimation) with $2,975$ training and $500$ testing images.
	\end{enumerate*}
	
	\paragraph{Baselines}
	The proposed DB-MTL is compared with a number of MTL baselines,
	including
	\begin{enumerate*}[(i), series = tobecont, itemjoin = \quad]
		\item equal weighting (EW) \cite{zhang2021survey}; 
		\item GLS \cite{chennupati2019multinet}, which minimizes the geometric mean loss $\sqrt[T]{\prod_{t=1}^T\ell_t(\hD_t; \vtheta, \vpsi_t)}$;
		\item RLW \cite{linreasonable}, in which the task weights are sampled from the standard normal distribution;
		\item \textit{loss balancing} methods including UW \cite{kendall2018multi}, DWA \cite{ljd19}, IMTL-L \cite{liu2021imtl}, and IGBv2 \cite{dai2023improvable};
		\item \textit{gradient balancing} methods including MGDA \cite{sk18}, GradNorm \cite{chen2018gradnorm}, PCGrad \cite{yu2020gradient}, GradDrop \cite{chen2020just}, GradVac \cite{wang2021gradient}, IMTL-G \cite{liu2021imtl}, CAGrad \cite{liu2021conflict}, MTAdam \cite{malkiel2021mtadam},
		Nash-MTL \cite{navon2022multi}, MetaBalance \cite{he2022metabalance},
		MoCo \cite{fernando2023mitigating}, and Aligned-MTL
		\cite{senushkin2023independent}; and
		\item 
		\textit{hybrid balancing} method IMTL \cite{liu2021imtl}. 
	\end{enumerate*}
	For comparison, we also include the \textit{single-task learning} (STL) baseline, which learns each task separately.
	
	All methods are implemented based on the open-source \texttt{LibMTL} library \cite{lin2022libmtl}. For all MTL methods,
	the hard-parameter sharing (\textit{HPS}) pattern \cite{Caruana93} is used, which consists of a task-sharing feature encoder and $T$ task-specific heads. For the proposed DB-MTL, following MoCo \cite{fernando2023mitigating}, we perform grid search for $\beta$ over $\{0.1, 0.5, 0.9, \frac{0.1}{k^{0.5}}, \frac{0.5}{k^{0.5}}, \frac{0.9}{k^{0.5}}\}$ for each dataset, where $k$ is the number of iterations.
	
	\paragraph{Implementation Details}
	Following RLW \cite{linreasonable}, we use the \textit{DeepLabV3+} network \cite{ChenZPSA18}, which contains a \textit{ResNet-50} network with dilated convolutions pre-trained on the \textit{ImageNet} dataset \cite{deng2009imagenet} as the shared encoder and the \textit{Atrous Spatial Pyramid Pooling} \cite{ChenZPSA18} module as task-specific head.  
	We train the model for $200$ epochs by using the Adam optimizer \cite{kingma2015adam} with learning rate $10^{-4}$ and weight decay $10^{-5}$. The learning rate is halved to $5\times10^{-5}$ after $100$ epochs.
	{The cross-entropy loss $\ell_{seg} = -\frac{1}{N \times H \times W}\sum_{n=1}^{N}\sum_{i=1}^{H \times W}\sum_{c=1}^{C} y_{n,i,c} \log(\hat{y}_{n,i,c})$, $L_1$ loss $\ell_{depth} = \frac{1}{N \times H \times W}\sum_{n=1}^{N}\sum_{i=1}^{H \times W}|d_{n,i} - \hat{d}_{n,i}|$, and cosine loss $\ell_{normal} = \frac{1}{N \times H \times W}\sum_{n=1}^{N}\sum_{i=1}^{H \times W}(1 - \frac{\mathbf{n}_{n,i} \cdot \hat{\mathbf{n}}_{n,i}}{||\mathbf{n}_{n,i}|| \cdot ||\hat{\mathbf{n}}_{n,i}||})$ are used as the loss functions of the semantic segmentation, depth estimation, and surface normal prediction tasks, respectively, where $N$ is the batch size, $H$ and $W$ are the height and width of the image, $y_{n,i,c}$ and $\hat{y}_{n,i,c}$ are the ground truth label and predicted probability for pixel $i$ in image $n$ and class $c$, $d_{n,i}$ and $\hat{d}_{n,i}$ are the ground truth and predicted depth values for pixel $i$ in image $n$, and $\mathbf{n}_{n,i}$ and $\hat{\mathbf{n}}_{n,i}$ are the ground truth and predicted normal vectors for pixel $i$ in image $n$.}
	For \textit{NYUv2}, the images are resized to $288\times 384$, and the batch size is $8$. For \textit{Cityscapes}, the images 
	are resized to $128\times 256$, and the batch size is $64$. Each experiment is repeated three times.
	
	\begin{table*}[!t]
		\centering
		\caption{Performance on \textit{NYUv2} with $3$ tasks. $\uparrow (\downarrow)$ means the higher (lower) the result, the better the performance. The best and second best results are marked in \textbf{bold} and \underline{underline}, respectively.}
		\label{tbl:mtl-nyu_dmtl}
		\resizebox{\textwidth}{!}{
			\begin{tabular}{lcccccccccc}
				\toprule
				\multicolumn{1}{l}{\multirow{4}{*}{\textbf{}}} & \multicolumn{2}{c}{\textbf{Segmentation}} & \multicolumn{2}{c}{\textbf{Depth Estimation}} & \multicolumn{5}{c}{\textbf{Surface Normal Prediction}} & 
				\multirow{4.5}{*}{\bm{$\Delta_{\mathrm{p}}$}${\uparrow}$}\\
				\cmidrule(lr){2-3} \cmidrule(lr){4-5} \cmidrule(lr){6-10}
				& \multicolumn{1}{c}{\multirow{2.5}{*}{\textbf{mIoU${\uparrow}$}}} &  \multicolumn{1}{c}{\multirow{2.5}{*}{\textbf{PAcc$\uparrow$}}} &  \multicolumn{1}{c}{\multirow{2.5}{*}{\textbf{AErr$\downarrow$}}} &  \multicolumn{1}{c}{\multirow{2.5}{*}{\textbf{RErr$\downarrow$}}} & \multicolumn{2}{c}{\textbf{Angle Distance}} & \multicolumn{3}{c}{\textbf{Within $t^{\circ}$}} \\ \cmidrule(lr){6-7} \cmidrule(lr){8-10} & & & & & \multicolumn{1}{c}{\textbf{Mean$\downarrow$}} & \multicolumn{1}{c}{\textbf{MED$\downarrow$}}  & \multicolumn{1}{c}{\textbf{11.25$\uparrow$}} & \multicolumn{1}{c}{\textbf{22.5$\uparrow$}} & \multicolumn{1}{c}{\textbf{30$\uparrow$}}  \\
				\midrule
				STL & $53.50$ & $75.39$ & $0.3926$ & $0.1605$ & $\underline{21.99}$ & $\textbf{15.16}$ & $\textbf{39.04}$ & $\textbf{65.00}$ & $\underline{75.16}$ & $0.00$ \\
				\midrule
				EW & $53.93$ & $75.53$ & $0.3825$ & $0.1577$ & $23.57$ & $17.01$ & $35.04$ & $60.99$ & $72.05$ & $\textcolor{teal}{-1.78}_{\pm0.45}$\\
				GLS & $\underline{54.59}$ & $\textbf{76.06}$ & $\underline{0.3785}$ & $\textbf{0.1555}$ & $22.71$ & $16.07$ & $36.89$ & $63.11$ & $73.81$ & $\underline{\textcolor{purple}{+0.30}}_{\pm0.30}$\\
				RLW & $54.04$ & $75.58$ & $0.3827$ & $0.1588$ & $23.07$ & $16.49$ & $36.12$ & $62.08$ & $72.94$ & $\textcolor{teal}{-1.10}_{\pm0.40}$ \\
				\midrule
				UW & ${54.29}$ & $75.64$ & $0.3815$ & $0.1583$ & $23.48$ & $16.92$ & $35.26$ & $61.17$ & $72.21$ & $\textcolor{teal}{-1.52}_{\pm0.39}$\\
				DWA & $54.06$ & $75.64$ & $0.3820$ & $0.1564$ & $23.70$ & $17.11$ & $34.90$ & $60.74$ & $71.81$ & $\textcolor{teal}{-1.71}_{\pm0.25}$\\
				IMTL-L & $53.89$ & $75.54$ & $0.3834$ & $0.1591$ & $23.54$ & $16.98$ & $35.09$ & $61.06$ & $72.12$ & $\textcolor{teal}{-1.92}_{\pm0.25}$\\
				IGBv2 & $\textbf{54.61}$ & $\underline{76.00}$ & $0.3817$ & $0.1576$ & $22.68$ & $15.98$ & $37.14$ & $63.25$ & $73.87$ & $\textcolor{purple}{+0.05}_{\pm0.29}$\\
				\midrule
				MGDA & $53.52$ & $74.76$ & $0.3852$ & $0.1566$ & $22.74$ & $16.00$ & $37.12$ & $63.22$ & $73.84$ & $\textcolor{teal}{-0.64}_{\pm0.25}$\\
				GradNorm & $53.91$ & $75.38$ & $0.3842$ & $0.1571$ & $23.17$ & $16.62$ & $35.80$ & $61.90$ & $72.84$ & $\textcolor{teal}{-1.24}_{\pm0.15}$\\
				PCGrad & $53.94$ & $75.62$ & $0.3804$ & $0.1578$ & $23.52$ & $16.93$ & $35.19$ & $61.17$ & $72.19$ & $\textcolor{teal}{-1.57}_{\pm0.44}$\\
				GradDrop & $53.73$ & $75.54$ & $0.3837$ & $0.1580$ & $23.54$ & $16.96$ & $35.17$ & $61.06$ & $72.07$ & $\textcolor{teal}{-1.85}_{\pm0.39}$ \\
				GradVac & $54.21$ & ${75.67}$ & $0.3859$ & $0.1583$ & $23.58$ & $16.91$ & $35.34$ & $61.15$ & $72.10$ & $\textcolor{teal}{-1.75}_{\pm0.39}$ \\
				IMTL-G & $53.01$ & $75.04$ & $0.3888$ & $0.1603$ & $23.08$ & $16.43$ & $36.24$ & $62.23$ & $73.06$ & $\textcolor{teal}{-1.89}_{\pm0.54}$ \\
				CAGrad & $53.97$ & $75.54$ & $0.3885$ & $0.1588$ & $22.47$ & $15.71$ & $37.77$ & $63.82$ & $74.30$ & $\textcolor{teal}{-0.27}_{\pm0.35}$\\
				MTAdam & $52.67$ & $74.86$ & $0.3873$ & $0.1583$ & $23.26$ & $16.55$ & $36.00$ & $61.92$ & $72.74$ & $\textcolor{teal}{-1.97}_{\pm0.23}$\\
				Nash-MTL & $53.41$ & $74.95$ & $0.3867$ & $0.1612$ & $22.57$ & $15.94$ & $37.30$ & $63.40$ & $74.09$ & $\textcolor{teal}{-1.01}_{\pm0.13}$\\
				MetaBalance & $53.92$ & $75.57$ & $0.3901$ & $0.1594$ & $22.85$ & $16.16$ & $36.72$ & $62.91$ & $73.62$ & $\textcolor{teal}{-1.06}_{\pm0.17}$\\
				MoCo & $52.25$ & $74.56$ & $0.3920$ & $0.1622$ & $22.82$ & $16.24$ & $36.58$ & $62.72$ & $73.49$ & $\textcolor{teal}{-2.25}_{\pm0.51}$\\
				Aligned-MTL & $52.94$ & $75.00$ & $0.3884$ & $0.1570$ & $22.65$ & $16.07$ & $36.88$ & $63.18$ & $73.94$ & $\textcolor{teal}{-0.98}_{\pm0.56}$\\
				\midrule
				IMTL & $53.63$ & $75.44$ & $0.3868$ & $0.1592$ & $22.58$ & $15.85$ & $37.44$ & $63.52$ & $74.09$ & $\textcolor{teal}{-0.57}_{\pm0.24}$\\
				% \midrule
				DB-MTL (\textbf{ours}) & $53.92$ & $75.60$ & $\textbf{0.3768}$ & $\underline{0.1557}$ & $\textbf{21.97}$ & $\underline{15.37}$ & $\underline{38.43}$ & ${64.81}$ & $\textbf{75.24}$ & $\textcolor{purple}{\textbf{+1.15}}_{\pm0.16}$ \\
				\bottomrule
		\end{tabular}}
	\end{table*}
	
	\begin{table}[!t]
		\centering
		\caption{Performance on \textit{Cityscapes} with $2$ tasks. $\uparrow (\downarrow)$ indicates that the higher (lower) the result, the better the performance. The best and second best results are highlighted in \textbf{bold} and \underline{underline}, respectively.}
		\label{tbl:mtl-Cityscapes}
		\resizebox{0.8\linewidth}{!}{
			\begin{tabular}{lccccc}
				\toprule
				& \multicolumn{2}{c}{\textbf{Segmentation}} & \multicolumn{2}{c}{\textbf{Depth Estimation}}& 
				\multirow{2.5}{*}{\bm{$\Delta_{\mathrm{p}}$}${\uparrow}$}\\
				\cmidrule(r){2-3} \cmidrule(lr){4-5}
				& \multicolumn{1}{c}{\textbf{mIoU${\uparrow}$}} &  \multicolumn{1}{c}{\textbf{PAcc$\uparrow$}} &  \multicolumn{1}{c}{\textbf{AErr$\downarrow$}} & \multicolumn{1}{c}{\textbf{RErr$\downarrow$}} \\
				\midrule
				STL & $69.06$ & $91.54$ & $0.01282$ & $\underline{43.53}$ & $\underline{0.00}$\\
				\midrule
				EW & $68.93$ & $91.58$ & $0.01315$ & $45.90$ & $\textcolor{teal}{-2.05}_{\pm0.56}$\\
				GLS & $68.69$ & $91.45$ & $0.01280$ & $44.13$ & $\textcolor{teal}{-0.39}_{\pm1.06}$ \\
				RLW & $69.03$ & $91.57$ & $0.01343$ & $44.77$ & $\textcolor{teal}{-1.91}_{\pm0.21}$\\
				\midrule
				UW & $69.03$ & $\underline{91.61}$ & $0.01338$ & $45.89$ & $\textcolor{teal}{-2.45}_{\pm0.68}$\\
				DWA & $68.97$ & $91.58$ & $0.01350$ & $45.10$ & $\textcolor{teal}{-2.24}_{\pm0.28}$\\
				IMTL-L & $68.98$ & $91.59$ & $0.01340$ & $45.32$ & $\textcolor{teal}{-2.15}_{\pm0.88}$\\
				IGBv2 & $68.44$ & $91.31$ & $0.01290$ & $45.03$ & $\textcolor{teal}{-1.31}_{\pm0.61}$\\
				\midrule
				MGDA & $69.05$ & $91.53$ & ${0.01280}$ & $44.07$ & $\textcolor{teal}{-0.19}_{\pm0.30}$\\
				GradNorm & $68.97$ & $91.60$ & $0.01320$ & $44.88$ & $\textcolor{teal}{-1.55}_{\pm0.70}$ \\
				PCGrad & $68.95$ & $91.58$ & $0.01342$ & $45.54$ & $\textcolor{teal}{-2.36}_{\pm1.17}$\\
				GradDrop & $68.85$ & $91.54$ & $0.01354$ & $44.49$ & $\textcolor{teal}{-2.02}_{\pm0.74}$ \\
				GradVac & $68.98$ & $91.58$ & $0.01322$ & $46.43$ & $\textcolor{teal}{-2.45}_{\pm0.54}$\\
				IMTL-G & $69.04$ & $91.54$ & $0.01280$ & $44.30$ & $\textcolor{teal}{-0.46}_{\pm0.67}$\\
				CAGrad & $68.95$ & $91.60$ & $0.01281$ & $45.04$ & $\textcolor{teal}{-0.87}_{\pm0.88}$\\
				MTAdam & $68.43$ & $91.26$ & $0.01340$ & $45.62$ & $\textcolor{teal}{-2.74}_{\pm0.20}$\\
				Nash-MTL & $68.88$ & $91.52$ & $\textbf{0.01265}$ & $45.92$ & $\textcolor{teal}{-1.11}_{\pm0.21}$\\
				MetaBalance & $69.02$ & $91.56$ & $\underline{0.01270}$ & $45.91$ & $\textcolor{teal}{-1.18}_{\pm0.58}$\\
				MoCo & $\textbf{69.62}$ & $\textbf{91.76}$ & $0.01360$ & $45.50$ & $\textcolor{teal}{-2.40}_{\pm1.50}$ \\
				Aligned-MTL & $69.00$ & $91.59$ & $0.01270$ & $44.54$ & $\textcolor{teal}{-0.43}_{\pm0.44}$\\
				\midrule
				IMTL & $69.07$ & $91.55$ & ${0.01280}$ & $44.06$ & $\textcolor{teal}{-0.32}_{\pm0.10}$\\
				DB-MTL (\textbf{ours}) & $\underline{69.17}$ & $91.56$ & ${0.01280}$ & $\textbf{43.46}$ & $\textcolor{purple}{\textbf{+0.20}}_{\pm0.40}$ \\
				\bottomrule
		\end{tabular}}
	\end{table}
	
	\begin{table*}[!t]
		\centering
		\caption{Performance (MAE) on \textit{QM9} with $11$ tasks. $\uparrow (\downarrow)$ indicates that the higher (lower) the result, the better the performance.  The best and second best results are highlighted in \textbf{bold} and \underline{underline}, respectively.} 
		\label{tbl:qm9}
		\resizebox{\textwidth}{!}{
			\begin{tabular}{lcccccccccccc}
				\toprule
				& \multicolumn{1}{c}{$\bm{\mu}$} & \multicolumn{1}{c}{$\bm{\alpha}$} & \multicolumn{1}{c}{$\bm{\epsilon_{\mathrm{HOMO}}}$} & \multicolumn{1}{c}{$\bm{\epsilon_{\mathrm{LUMO}}}$} & \multicolumn{1}{c}{$\bm{\langle R^2 \rangle}$} & \multicolumn{1}{c}{\textbf{ZPVE}} & \multicolumn{1}{c}{$\bm{U_0}$} & \multicolumn{1}{c}{$\bm{U}$} & \multicolumn{1}{c}{$\bm{H}$} & \multicolumn{1}{c}{$\bm{G}$} & \multicolumn{1}{c}{$\bm{c_v}$} & \multirow{1}{*}{\bm{$\Delta_{\mathrm{p}}$}${\uparrow}$} \\
				%\cmidrule(lr){2-12}
				%& \multicolumn{11}{c}{\textbf{MAE}${\downarrow}$} \\
				\midrule
				STL & $\textbf{0.062}$ & $\textbf{0.192}$ & $\textbf{58.82}$ & $\textbf{51.95}$ & $\textbf{0.529}$ & $4.52$ & $63.69$ & $60.83$ & $68.33$ & $60.31$ & $\textbf{0.069}$ & $\textbf{0.00}$\\
				\midrule
				EW & ${0.096}$ & $0.286$ & $67.46$ & ${82.80}$ & $4.655$ & $12.4$ & $128.3$ & $128.8$ & $129.2$ & $125.6$ & $0.116$ & $\textcolor{teal}{-146.3}_{\pm7.86}$\\
				GLS & $0.332$ & $0.340$ & $143.1$ & $131.5$ & $1.023$ & $\textbf{4.45}$ & $\textbf{53.35}$ & $\textbf{53.79}$ & $\textbf{53.78}$ & $\textbf{53.34}$ & $0.111$ & $\textcolor{teal}{-81.16}_{\pm15.5}$\\
				RLW & $0.112$ & $0.331$ & $74.59$ & $90.48$ & $6.015$ & $15.6$ & $156.0$ & $156.8$ & $157.3$ & $151.6$ & $0.133$ & $\textcolor{teal}{-200.9}_{\pm13.4}$\\
				\midrule
				UW & $0.336$ & $0.382$ & $155.1$ & $144.3$ & $0.965$ & $4.58$ & $61.41$ & $61.79$ & $61.83$ & $61.40$ & $0.116$ & $\textcolor{teal}{-92.35}_{\pm13.9}$\\
				DWA & $0.103$ & $0.311$ & $71.55$ & $87.21$ & $4.954$ & $13.1$ & $134.9$ & $135.8$ & $136.3$ & $132.0$ & $0.121$ & $\textcolor{teal}{-160.9}_{\pm16.7}$\\
				IMTL-L & $0.277$ & $0.355$ & $150.1$ & $135.2$ & $\underline{0.946}$ & $\underline{4.46}$ & $\underline{58.08}$ & $\underline{58.43}$ & $\underline{58.46}$ & $\underline{58.06}$ & $0.110$ & $\textcolor{teal}{-77.06}_{\pm11.1}$\\
				IGBv2 & $0.235$ & $0.377$ & $132.3$ & $139.9$ & $2.214$ & $5.90$ & $64.55$ & $65.06$ & $65.12$ & $64.28$ & $0.121$ & $\textcolor{teal}{-99.86}_{\pm10.4}$\\
				\midrule
				MGDA & $0.181$ & $0.325$ & $118.6$ & $92.45$ & $2.411$ & $5.55$ & $103.7$ & $104.2$ & $104.4$ & $103.7$ & $0.110$ & $\textcolor{teal}{-103.0}_{\pm8.62}$\\
				GradNorm & $0.114$ & $0.341$ & $\underline{67.17}$ & $84.66$ & $7.079$ & $14.6$ & $173.2$ & $173.8$ & $174.4$ & $168.9$ & $0.147$ & $\textcolor{teal}{-227.5}_{\pm1.85}$\\
				PCGrad & $0.104$ & $0.293$ & $75.29$ & $88.99$ & $3.695$ & $8.67$ & $115.6$ & $116.0$ & $116.2$ & $113.8$ & $0.109$ & $\textcolor{teal}{-117.8}_{\pm3.97}$\\
				GradDrop & $0.114$ & $0.349$ & $75.94$ & $94.62$ & $5.315$ & $15.8$ & $155.2$ & $156.1$ & $156.6$ & $151.9$ & $0.136$ & $\textcolor{teal}{-191.4}_{\pm9.62}$\\
				GradVac & $0.100$ & $0.299$ & $68.94$ & $84.14$ & $4.833$ & $12.5$ & $127.3$ & $127.8$ & $128.1$ & $124.7$ & $0.117$ & $\textcolor{teal}{-150.7}_{\pm7.41}$\\
				IMTL-G & $0.670$ & $0.978$ & $220.7$ & $249.7$ & $19.48$ & $55.6$ & $1109$ & $1117$ & $1123$ & $1043$ & $0.392$ & $\textcolor{teal}{-1250}_{\pm90.9}$\\
				CAGrad & $0.107$ & $0.296$ & $75.43$ & $88.59$ & $2.944$ & $6.12$ & $93.09$ & $93.68$ & $93.85$ & $92.32$ & $0.106$ & $\textcolor{teal}{-87.25}_{\pm1.51}$\\
				MTAdam & $0.593$ & $1.352$ & $232.3$ & $419.0$ & $24.31$ & $69.7$ & $1060$ & $1067$ & $1070$ & $1007$ & $0.627$ & $\textcolor{teal}{-1403}_{\pm203}$\\
				Nash-MTL & $0.115$ & $\underline{0.263}$ & $85.54$ & $86.62$ & $2.549$ & $5.85$ & $83.49$ & $83.88$ & $84.05$ & $82.96$ & $\underline{0.097}$ & $\textcolor{teal}{-73.92}_{\pm2.12}$\\
				MetaBalance & $\underline{0.090}$ & $0.277$ & $70.50$ & $\underline{78.43}$ & $4.192$ & $11.2$ & $113.7$ & $114.2$ & $114.5$ & $111.7$ & $0.110$ & $\textcolor{teal}{-125.1}_{\pm7.98}$\\
				MoCo & $0.489$ & $1.096$ & $189.5$ & $247.3$ & $34.33$ & $64.5$ & $754.6$ & $760.1$ & $761.6$ & $720.3$ & $0.522$ & $\textcolor{teal}{-1314}_{\pm65.2}$\\
				Aligned-MTL & $0.123$ & $0.295$ & $98.07$ & $94.56$ & $2.397$ & $5.90$ & $86.42$ & $87.42$ & $87.19$ & $86.75$ & $0.106$ & $\textcolor{teal}{-80.58}_{\pm4.18}$\\
				\midrule
				IMTL & $0.138$ & $0.344$ & $106.1$ & $102.9$ & $2.595$ & $7.84$ & $102.5$ & $103.0$ & $103.2$ & $100.8$ & $0.110$ & $\textcolor{teal}{-104.3}_{\pm11.7}$\\
				DB-MTL (\textbf{ours}) & $0.112$ & $0.264$ & $89.26$ & $86.59$ & $2.429$ & $5.41$ & $60.33$ & $60.78$ & $60.80$ & $60.59$ & $0.098$ & $\underline{\textcolor{teal}{-58.10}}_{\pm3.89}$\\
				\bottomrule
		\end{tabular}}
	\end{table*}
	
	\paragraph{Performance Evaluation} 
	Following DWA \cite{ljd19} and RLW \cite{linreasonable}, we use
	\begin{enumerate*}[(i), series = tobecont, itemjoin = \quad]
		\item the mean intersection over union (mIoU) and class-wise pixel accuracy (PAcc) for semantic segmentation;
		\item relative error (RErr) and absolute error (AErr) for depth estimation; 
		\item mean and median angle errors, and percentage of normals within $t^{\circ}$ (where $t=11.25, 22.5, 30$) for surface normal prediction.
	\end{enumerate*}
	Following \cite{maninis2019attentive, vandenhende2021multi, linreasonable}, we report the relative performance improvement of an MTL method $\hA$ over STL, averaged over all the metrics above, i.e.,
	\begin{equation} \label{eq:delta_p}
		\Delta_{\mathrm{p}} (\hA)=\frac{1}{T}\sum_{t=1}^{T}\Delta_{\mathrm{p},t} (\hA), 
	\end{equation} 
	where $T$ is the number of tasks and
	\begin{equation*} 
		\Delta_{\mathrm{p},t} (\hA) = 100\%\times\frac{1}{N_{t}}\sum_{i=1}^{N_{t}}(-1)^{s_{t, i}}\frac{M_{t, i}^\hA-M_{t, i}^{\text{STL}}}{M_{t, i}^{\text{STL}}},
	\end{equation*} 
	where $N_t$ is the number of metrics for task $t$, $M_{t,i}^\hA$ is the $i$th metric value of method $\hA$ on task $t$, and $s_{t, i}$ is $0$ if a larger value indicates better performance for the $i$th metric on task $t$, and 1 otherwise.
	
	\paragraph{Performance Results} 
	Table \ref{tbl:mtl-nyu_dmtl} shows the results on \textit{NYUv2}. As can be seen, the proposed DB-MTL performs the best in terms of average $\Delta_{\mathrm{p}}$. Note that most of the MTL baselines perform better than STL on semantic segmentation and depth estimation, but have a large drop on the surface normal prediction task, suffering from the task balancing problem. 
	Only the proposed DB-MTL has comparable performance with STL on the surface normal prediction task and maintains superiority on the other tasks.
	
	Table \ref{tbl:mtl-Cityscapes} shows the results on \textit{Cityscapes}. As can be seen, DB-MTL again achieves the best in terms of average $\Delta_{\mathrm{p}}$. Note that all MTL baselines perform worse than STL in terms of average $\Delta_{\mathrm{p}}$ and only the proposed DB-MTL outperforms STL on all tasks. 
	
	\subsection{Evaluation on Molecular Property Prediction}
	\label{sec:mo}
	
	\paragraph{Dataset} 
	Following Nash-MTL \cite{navon2022multi}, we use the \textit{QM9}
	\cite{ramakrishnan2014quantum} dataset, which is for molecular property prediction with $11$ tasks. Each task performs regression on one property. We use the same split as in Nash-MTL \cite{navon2022multi}: $110, 000$ for training, $10,000$ for
	validation, and $10,000$ for testing.
	
	\begin{table*}[!t]
		\centering
		\caption{Classification accuracy (\%) on \textit{Office-31} and \textit{Office-Home}.  $\uparrow$ indicates that the higher the result, the better the performance.  The best and second best results are highlighted in \textbf{bold} and \underline{underline}, respectively. Results of MoCo are from \cite{fernando2023mitigating}.} 
		\label{tbl:mtl-31-home}
		\resizebox{\linewidth}{!}{
			\begin{tabular}{lcccccc@{\hspace{.2cm}}cccccc}
				\toprule
				& \multicolumn{5}{c}{\textit{Office-31}} & & \multicolumn{6}{c}{\textit{Office-Home}}\\
				\cmidrule(r){2-6}\cmidrule{8-13}
				&\textbf{Amazon} & \multicolumn{1}{c}{\textbf{DSLR}} & \multicolumn{1}{c}{\textbf{Webcam}} & \multirow{1}{*}{{\textbf{Avg}$\uparrow$}} & \multirow{1}{*}{\bm{$\Delta_{\mathrm{p}}$}$\uparrow$} & &\textbf{Artistic} & \multicolumn{1}{c}{\textbf{Clipart}} & \multicolumn{1}{c}{\textbf{Product}} & \multicolumn{1}{c}{\textbf{Real}} & \multirow{1}{*}{{\textbf{Avg}$\uparrow$}} & \multirow{1}{*}{\bm{$\Delta_{\mathrm{p}}$}$\uparrow$}\\
				\midrule
				STL & $\textbf{86.61}$ & $95.63$ & $96.85$ & $93.03$ & 0.00 & & $65.59$ & $\textbf{79.60}$ & $\textbf{90.47}$ & $80.00$ & $78.91$ & $0.00$\\
				\midrule
				EW & $83.53$ & $97.27$ & $96.85$ & $92.55_{\pm0.62}$ & $\textcolor{teal}{-0.61}_{\pm0.67}$ & & $65.34$ & $78.04$ & $89.80$ & $79.50$ & $78.17_{\pm0.37}$ & $\textcolor{teal}{-0.92}_{\pm0.59}$\\
				GLS & $82.84$ & $95.62$ & $96.29$ & $91.59_{\pm0.58}$ & $\textcolor{teal}{-1.63}_{\pm0.61}$ & & $64.51$ & $76.85$ & $89.83$ & $79.56$ & $77.69_{\pm0.27}$ & $\textcolor{teal}{-1.58}_{\pm0.46}$\\
				RLW & $83.82$ & $96.99$ & $96.85$ & $92.55_{\pm0.89}$ & $\textcolor{teal}{-0.59}_{\pm0.95}$ & & $64.96$ & $78.19$ & $89.48$ & $\textbf{80.11}$ & $78.18_{\pm0.12}$ & $\textcolor{teal}{-0.92}_{\pm0.14}$\\
				\midrule
				UW & $83.82$ & $97.27$ & $96.67$ & $92.58_{\pm0.84}$ & $\textcolor{teal}{-0.56}_{\pm0.90}$ & & $65.97$ & $77.65$ & $89.41$ & $79.28$ & $78.08_{\pm0.30}$ & $\textcolor{teal}{-0.98}_{\pm0.46}$\\
				DWA & $83.87$ & $96.99$ & $96.48$ & $92.45_{\pm0.56}$ & $\textcolor{teal}{-0.70}_{\pm0.62}$ & & $65.27$ & $77.64$ & $89.05$ & $79.56$ & $77.88_{\pm0.28}$ & $\textcolor{teal}{-1.26}_{\pm0.49}$\\
				IMTL-L & $84.04$ & $96.99$ & $96.48$ & $92.50_{\pm0.52}$ & $\textcolor{teal}{-0.63}_{\pm0.58}$ & & $65.90$ & $77.28$ & $89.37$ & $79.38$ & $77.98_{\pm0.38}$ & $\textcolor{teal}{-1.10}_{\pm0.61}$\\
				IGBv2 & $84.52$ & $\underline{98.36}$ & ${98.05}$ & $\underline{93.64}_{\pm0.26}$ & $\underline{\textcolor{purple}{+0.56}}_{\pm0.25}$ & & $65.59$ & $77.57$ & $89.79$ & $78.73$ & $77.92_{\pm0.21}$ & $\textcolor{teal}{-1.21}_{\pm0.22}$\\
				\midrule
				MGDA & $\underline{85.47}$ & $95.90$ & $97.03$ & $92.80_{\pm0.14}$ & $\textcolor{teal}{-0.27}_{\pm0.15}$ & & $64.19$ & $77.60$ & $89.58$ & $79.31$ & $77.67_{\pm0.20}$ & $\textcolor{teal}{-1.61}_{\pm0.34}$\\
				GradNorm & $83.58$ & $97.26$ & $96.85$ & $92.56_{\pm0.87}$ & $\textcolor{teal}{-0.59}_{\pm0.94}$ & & $66.28$ & $77.86$ & $88.66$ & $79.60$ & $78.10_{\pm0.63}$ & $\textcolor{teal}{-0.90}_{\pm0.93}$\\
				PCGrad & $83.59$ & $96.99$ & $96.85$ & $92.48_{\pm0.53}$ & $\textcolor{teal}{-0.68}_{\pm0.57}$ & & $\underline{66.35}$ & $77.18$ & $88.95$ & $79.50$ & $77.99_{\pm0.19}$ & $\textcolor{teal}{-1.04}_{\pm0.32}$\\
				GradDrop & $84.33$ & $96.99$ & $96.30$ & $92.54_{\pm0.42}$ & $\textcolor{teal}{-0.59}_{\pm0.46}$ & & $63.57$ & $77.86$ & $89.23$ & $79.35$ & $77.50_{\pm0.23}$ & $\textcolor{teal}{-1.86}_{\pm0.24}$\\
				GradVac & $83.76$ & $97.27$ & $96.67$ & $92.57_{\pm0.73}$ & $\textcolor{teal}{-0.58}_{\pm0.78}$ & & $65.21$ & $77.43$ & $89.23$ & $78.95$ & $77.71_{\pm0.19}$ & $\textcolor{teal}{-1.49}_{\pm0.28}$\\
				IMTL-G & $83.41$ & $96.72$ & $96.48$ & $92.20_{\pm0.89}$ & $\textcolor{teal}{-0.97}_{\pm0.95}$ & & $64.70$ & $77.17$ & $89.61$ & $79.45$ & $77.98_{\pm0.38}$ & $\textcolor{teal}{-1.10}_{\pm0.61}$\\
				CAGrad & $83.65$ & $95.63$ & $96.85$ & $92.04_{\pm0.79}$ & $\textcolor{teal}{-1.14}_{\pm0.85}$ & & $64.01$ & $77.50$ & $89.65$ & $79.53$ & $77.73_{\pm0.16}$ & $\textcolor{teal}{-1.50}_{\pm0.29}$\\
				MTAdam & $85.52$ & $95.62$ & $96.29$ & $92.48_{\pm0.87}$ & $\textcolor{teal}{-0.60}_{\pm0.93}$ & & $62.23$ & $77.86$ & $88.73$ & $77.94$ & $76.69_{\pm0.65}$ & $\textcolor{teal}{-2.94}_{\pm0.85}$\\
				Nash-MTL & $85.01$ & ${97.54}$ & $97.41$ & $93.32_{\pm0.82}$ & $\textcolor{purple}{+0.24}_{\pm0.89}$ & & $66.29$ & $78.76$ & $90.04$ & $\textbf{80.11}$ & $\underline{78.80}_{\pm0.52}$ & $\underline{\textcolor{teal}{-0.08}}_{\pm0.69}$\\
				MetaBalance & $84.21$ & $95.90$ & $97.40$ & $92.50_{\pm0.28}$ & $\textcolor{teal}{-0.63}_{\pm0.30}$ & & $64.01$ & $77.50$ & $89.72$ & $79.24$ & $77.61_{\pm0.42}$ & $\textcolor{teal}{-1.70}_{\pm0.54}$\\
				MoCo & $84.33$ & ${97.54}$ & $\underline{98.33}$ & ${93.39}$ & - & & $63.38$ & $\underline{79.41}$ & $90.25$ & $78.70$ & $77.93$ & -\\
				Aligned-MTL & $83.36$ & $96.45$ & $97.04$ & $92.28_{\pm0.46}$ & $\textcolor{teal}{-0.90}_{\pm0.48}$ & & $64.33$ & $76.96$ & $89.87$ & $79.93$ & $77.77_{\pm0.70}$ & $\textcolor{teal}{-1.50}_{\pm0.89}$\\
				\midrule
				IMTL & $83.70$ & $96.44$ & $96.29$ & $92.14_{\pm0.85}$ & $\textcolor{teal}{-1.02}_{\pm0.92}$ & & $64.07$ & $76.85$ & $89.65$ & $79.81$ & $77.59_{\pm0.29}$ & $\textcolor{teal}{-1.72}_{\pm0.45}$\\
				% \midrule
				DB-MTL (\textbf{ours}) & $85.12$ & $\textbf{98.63}$ & $\textbf{98.51}$ & $\textbf{94.09}_{\pm0.19}$ & $\textcolor{purple}{\textbf{+1.05}}_{\pm0.20}$ & & $\textbf{67.42}$ & $77.89$ & $\underline{90.43}$ & $\underline{80.07}$ & $\textbf{78.95}_{\pm0.35}$ & $\textcolor{purple}{\textbf{+0.17}}_{\pm0.44}$\\ 
				\bottomrule
		\end{tabular}}
	\end{table*}
	
	\paragraph{Implementation Details}
	The experimental setups are the same with Nash-MTL \cite{navon2022multi}. Specifically, a graph neural network \cite{gilmer2017neural} is used as the shared encoder, and a linear layer is used as the task-specific head. 
	The targets of each task are normalized to have zero mean and unit standard deviation.
	The batch size and training epoch are set to $128$ and $300$, respectively. The Adam optimizer \cite{kingma2015adam} with the learning rate $0.001$ is used for training, and the ReduceLROnPlateau scheduler \cite{paszke2019pytorch} is used to reduce the learning rate once $\Delta_{\mathrm{p}}$ on the validation dataset stops improving. 
	{The mean squared error (MSE) $\ell_{mse} = \frac{1}{N}\sum_{n=1}^{N}(p_n - \hat{p}_n)^2$ is used as the loss function for each molecular property prediction task, where $N$ is the batch size, $p_n$ and $\hat{p}_n$ are the ground truth and predicted property values for sample $n$ respectively.} Mean absolute error (MAE) is used for performance evaluation. Each experiment is repeated three times.
	
	\paragraph{Performance Results} 
	Table \ref{tbl:qm9} shows each task's testing MAE and overall performance $\Delta_{\mathrm{p}}$ (Eq. (\ref{eq:delta_p})) on \textit{QM9}, using the same set of baselines as in Section~\ref{sec:nyu}. Note that \textit{QM9} is a challenging dataset in MTL and none of the MTL methods performs better than STL, as also observed in previous works \cite{johannes2020, navon2022multi}. 
	DB-MTL performs the best among all MTL methods and greatly improves over the second-best MTL method, Nash-MTL, in terms of average $\Delta_{\mathrm{p}}$.

	\subsection{Evaluation on Image Classification}
	\label{sec:image}
	
	\paragraph{Datasets} 
	Following RLW \cite{linreasonable} and MoCo \cite{fernando2023mitigating}, two image classification datasets are used:
	\begin{enumerate*}[(i), series = tobecont, itemjoin = \quad]
		\item \textit{Office-31} \cite{saenko2010adapting}, which contains $4,110$ images from three domains (tasks): Amazon, DSLR, and Webcam. Each task has 31 classes.
		\item \textit{Office-Home} \cite{venkateswara2017deep}, which contains $15,500$ images from four domains (tasks): artistic images, clipart, product images, and real-world images.
		Each task has $65$ object categories collected under office and home settings.
	\end{enumerate*}
	We use the commonly-used data split as in RLW \cite{linreasonable}: $60\%$ for training, $20\%$ for validation, and $20\%$ for testing. 
	
	\paragraph{Implementation Details}
	Following RLW \cite{linreasonable}, 
	a \textit{ResNet-18} \cite{he2016deep} pre-trained on the \textit{ImageNet} dataset \cite{deng2009imagenet} is used as a shared encoder, and a linear layer is used as a task-specific head. We resize the input image to $224\times 224$. 
	The batch size and number of training epochs are set to $64$ and $100$, respectively.
	The Adam optimizer \cite{kingma2015adam} with learning rate $10^{-4}$ and weight decay $10^{-5}$ is used. 
	{For each image classification task, the cross-entropy loss $\ell_{cls} = -\frac{1}{N}\sum_{n=1}^{N}\sum_{c=1}^{C} y_{n,c} \log(\hat{y}_{n,c})$ is used as the loss function, where $N$ is the batch size, $y_{n,c}$ is the ground truth label and $\hat{y}_{n,c}$ is the predicted probability for sample $n$ and class $c$.} Classification accuracy is used for evaluation. $\Delta_{\mathrm{p}}$ in Eq. (\ref{eq:delta_p}) is used as the overall performance metrics. Each experiment is repeated three times.
	
	\paragraph{Performance Results} 
	Table \ref{tbl:mtl-31-home} shows the results on \textit{Office-31} and \textit{Office-Home}, using the same set of baselines as in Section~\ref{sec:nyu}. 
	On \textit{Office-31}, DB-MTL achieves the top testing accuracy on the DSLR and Webcam tasks, and comparable performance on the Amazon task. On \textit{Office-Home}, DB-MTL ranks top two
	on the Artistic, Product, and Real tasks. On both datasets, DB-MTL achieves the best average testing accuracy and $\Delta_{\mathrm{p}}$, showing its effectiveness and demonstrating that balancing both loss scale and gradient magnitude is effective. 
	
	\begin{table*}[!t]
		\centering
		\caption{Effects of each component in DB-MTL on different datasets in terms of $\Delta_{\mathrm{p}}$ (Eq. (\ref{eq:delta_p})).} 
		\label{tbl:ablation}
		\resizebox{\linewidth}{!}{
			\begin{tabular}{cc|ccccc}
				\toprule
				loss-scale & gradient-magnitude & \multirow{2.5}{*}{\textit{NYUv2}} & \multirow{2.5}{*}{\textit{Cityscapes}} & \multirow{2.5}{*}{\textit{Office-31}} & \multirow{2.5}{*}{\textit{Office-Home}} & \multirow{2.5}{*}{\textit{QM9}}\\
				balancing & balancing \\
				\midrule
				\XSolidBrush & \XSolidBrush & ${-1.78}_{\pm0.45}$ & ${-2.05}_{\pm0.56}$ & ${-0.61}_{\pm0.67}$ & ${-0.92}_{\pm0.59}$ & ${-146.3}_{\pm7.86}$\\
				\Checkmark & \XSolidBrush & $+0.06_{\pm0.09}$ & $-0.38_{\pm0.39}$ & $+0.93_{\pm0.42}$ & $-0.73_{\pm0.95}$ & $-74.40_{\pm13.2}$\\
				\XSolidBrush & \Checkmark & $+0.76_{\pm0.25}$ & $+0.12_{\pm0.70}$ & $+0.01_{\pm0.39}$ & $-0.78_{\pm0.49}$ & $-65.73_{\pm2.86}$\\
				\Checkmark & \Checkmark & $\textbf{+1.15}_{\pm0.16}$ & $\textbf{+0.20}_{\pm0.40}$  & $\textbf{+1.05}_{\pm0.20}$ & $\textbf{+0.17}_{\pm0.44}$ & $\textbf{-58.10}_{\pm3.89}$\\
				\bottomrule
		\end{tabular}}
	\end{table*}
	
	\begin{figure*}[!t]
		\centering
		\includegraphics[width=\linewidth]{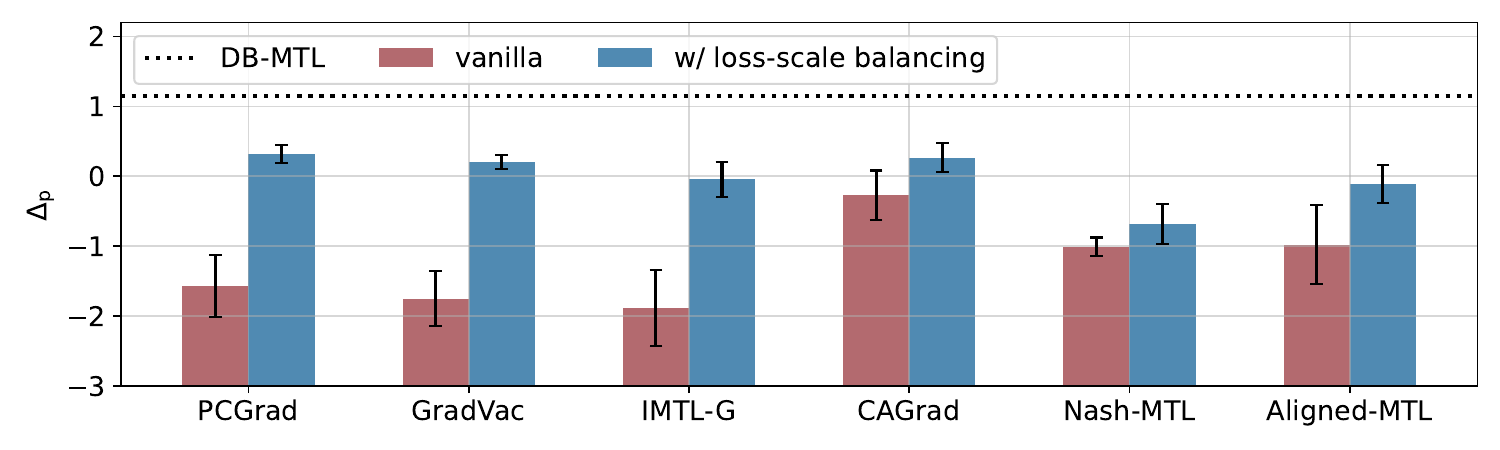} 
		%\vskip -0.15in
		\caption{Performance of existing gradient balancing methods with the loss-scale balancing method (i.e., logarithm transformation) on \textit{NYUv2}. ``vanilla'' stands for the original method.}
		\label{fig:log_grad}
	\end{figure*}
	
	\subsection{Effectiveness of Loss and Gradient Balancing Components}
	
	\paragraph{Ablation Study}
	\label{sec:ablation}
	DB-MTL has two components: loss-scale balancing (i.e., logarithm transformation) in Section \ref{sec:loss} and gradient-magnitude balancing in Section \ref{sec:si-g}. 
	In this experiment, we perform an ablation study on the effectiveness of each component. We consider the four combinations: 
	\begin{enumerate*}[(i), series = tobecont, itemjoin = \quad]
		\item use neither loss-scale nor gradient-magnitude balancing (i.e., the EW baseline); 
		\item use only loss-scale balancing;
		\item use only gradient-magnitude balancing;
		\item use both loss-scale and gradient-magnitude balancing (i.e., the proposed DB-MTL).
	\end{enumerate*}
	
	Table \ref{tbl:ablation} shows the $\Delta_{\mathrm{p}}$'s of the four combinations on five datasets (\textit{NYUv2}, \textit{Cityscapes}, \textit{Office-31}, \textit{Office-Home}, and \textit{QM9}). As can be seen, on all datasets, both components are beneficial to DB-MTL and combining them achieves the best performance.
	
	\paragraph{Effectiveness of Logarithm Transformation} \label{sec:log_grad}
	
	The logarithm transformation can also be used with other 
	gradient balancing methods. 
	We integrate it into
	PCGrad \cite{yu2020gradient}, GradVac
	\cite{wang2021gradient}, IMTL-G \cite{liu2021imtl}, CAGrad \cite{liu2021conflict}, Nash-MTL \cite{navon2022multi}, and Aligned-MTL \cite{senushkin2023independent}.
	The experiment is performed on \textit{NYUv2} using the setup in Section \ref{sec:nyu}. Figure \ref{fig:log_grad} shows the $\Delta_{\mathrm{p}}$ (Eq. \eqref{eq:delta_p}). 
	As can be seen, logarithm transformation is consistently
	beneficial	for these gradient balancing methods, showing the effectiveness of logarithm transformation. Moreover, DB-MTL still outperforms these gradient balancing baselines when they are combined with logarithm transformation, demonstrating the effectiveness of the proposed DB-MTL method.
	
\begin{figure*}[!t]
		\centering
		\begin{minipage}[b]{0.49\textwidth}
			\includegraphics[width=\linewidth]{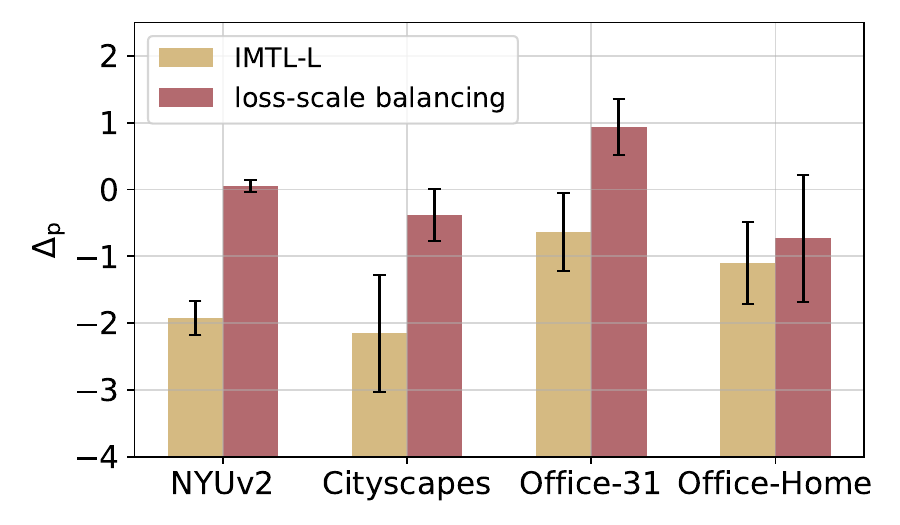}
			\caption{Comparison of IMTL-L \cite{liu2021imtl} and the loss-scale balancing method on four datasets.}
			\label{fig:log_per}
		\end{minipage}
		\hfill
		\begin{minipage}[b]{0.49\textwidth}
			\includegraphics[width=\linewidth]{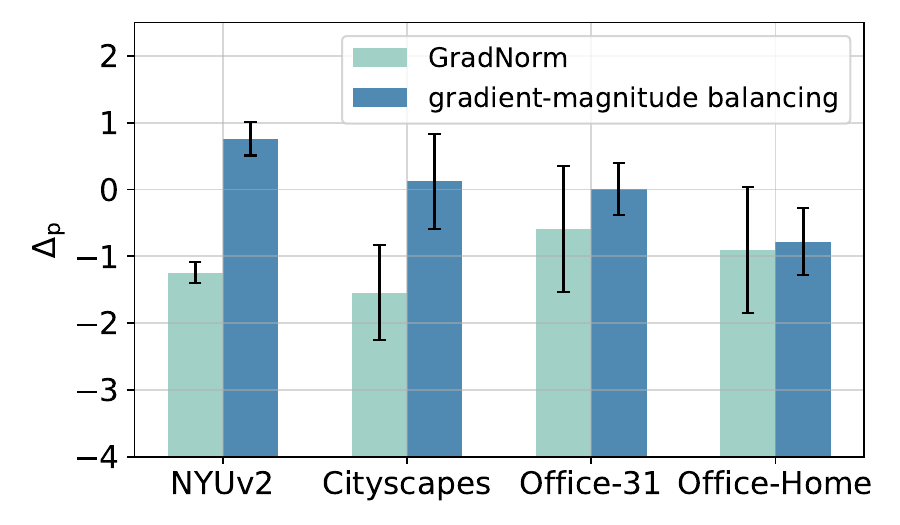}
			\caption{Comparison of GradNorm \cite{chen2018gradnorm} and the gradient-magnitude balancing method on four datasets.}
			\label{fig:grad_per}
		\end{minipage}
	\end{figure*}

	Further to the discussion in Section \ref{sec:loss}, we compare the loss-scale balancing method (i.e., using logarithm transformation only) with IMTL-L \cite{liu2021imtl} on four datasets (\textit{NYUv2}, \textit{Cityscapes}, \textit{Office-31}, and \textit{Office-Home}). As can be seen from Figure \ref{fig:log_per}, the logarithm transformation consistently outperforms IMTL-L in terms of average $\Delta_{\mathrm{p}}$ (Eq. (\ref{eq:delta_p})).
	
	\begin{figure*}
		\subfloat[\textit{Cross-stitch}.]{\label{fig:arch_cross}\includegraphics[width=0.5\linewidth]{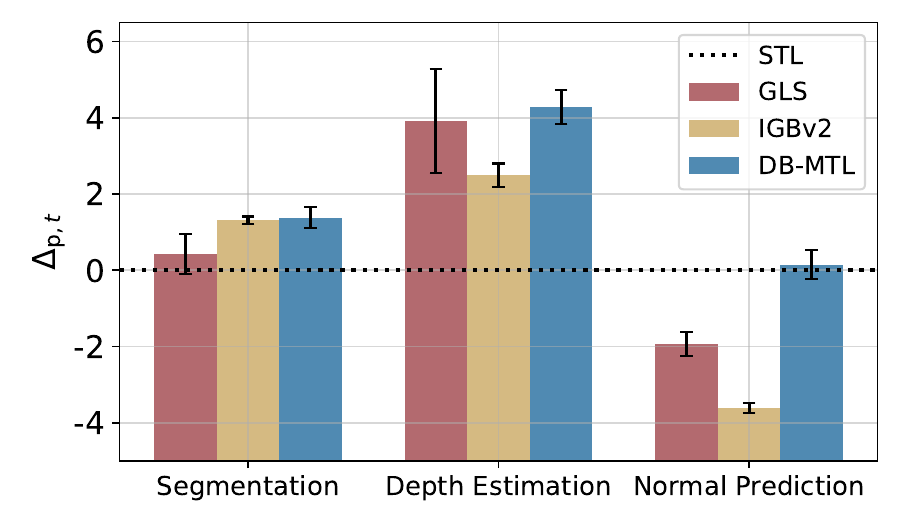}} 
		\hfill
		\subfloat[\textit{MTAN}.]{\label{fig:arch_mtan}\includegraphics[width=0.5\linewidth]{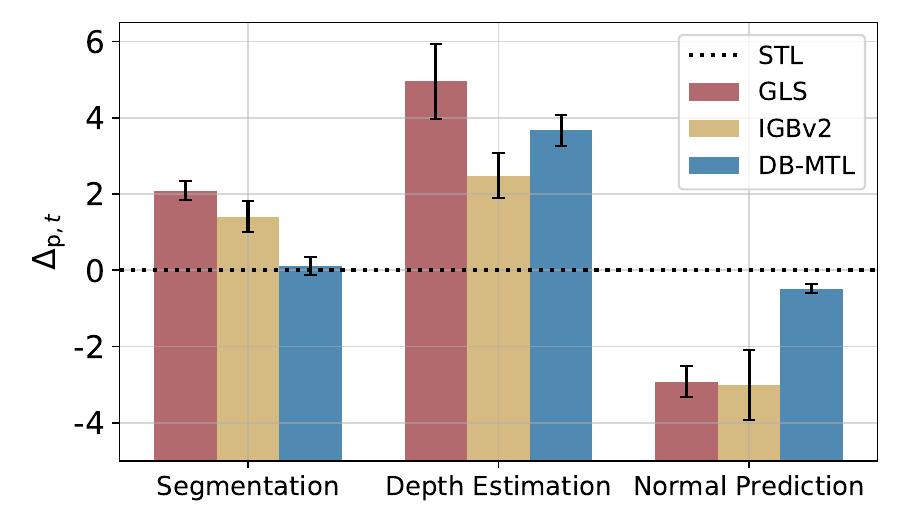}}
		\caption{Performance on \textit{NYUv2} for \textit{Cross-stitch} \cite{MisraSGH16} and \textit{MTAN} \cite{ljd19} architectures.}
		\label{fig:arch}
	\end{figure*}
	
	\begin{table*}[!t]
		\centering
		\caption{Performance on the \textit{NYUv2} dataset with \textit{SegNet} network. $\uparrow (\downarrow)$ indicates that the higher (lower) the result, the better the performance. The best and second best results are highlighted in \textbf{bold} and \underline{underline}, respectively. Superscripts $\sharp$, $\S$, $\ddag$, and $*$ denote the results are from \cite{liu2021conflict}, \cite{navon2022multi}, \cite{fernando2023mitigating}, and \cite{senushkin2023independent}, respectively.} 
		\label{tbl:mtl-nyu_segnet}
		\resizebox{\textwidth}{!}{
			\begin{tabular}{lcccccccccc}
				\toprule
				& \multicolumn{2}{c}{\textbf{Segmentation}} & \multicolumn{2}{c}{\textbf{Depth Estimation}} & \multicolumn{5}{c}{\textbf{Surface Normal Prediction}} & \multirow{4.5}{*}{\bm{$\Delta_{\mathrm{p}}$}$\uparrow$} \\
				\cmidrule(lr){2-3} \cmidrule(lr){4-5} \cmidrule(lr){6-10}
				& \multicolumn{1}{c}{\multirow{2.5}{*}{\textbf{mIoU${\uparrow}$}}} &  \multicolumn{1}{c}{\multirow{2.5}{*}{\textbf{PAcc$\uparrow$}}} &  \multicolumn{1}{c}{\multirow{2.5}{*}{\textbf{AErr$\downarrow$}}} &  \multicolumn{1}{c}{\multirow{2.5}{*}{\textbf{RErr$\downarrow$}}} & \multicolumn{2}{c}{\textbf{Angle Distance}} & \multicolumn{3}{c}{\textbf{Within $t^{\circ}$}} \\ 
				\cmidrule(lr){6-7} \cmidrule(lr){8-10} 
				& & & & & \multicolumn{1}{c}{\textbf{Mean$\downarrow$}} & \multicolumn{1}{c}{\textbf{MED$\downarrow$}}  & \multicolumn{1}{c}{\textbf{11.25$\uparrow$}} & \multicolumn{1}{c}{\textbf{22.5$\uparrow$}} & \multicolumn{1}{c}{\textbf{30$\uparrow$}}  \\
				\midrule
				STL$^\S$ & $38.30$ & $63.76$ & $0.6754$ & $0.2780$ & $25.01$ & $\underline{19.21}$ & $\underline{30.14}$ & $\underline{57.20}$ & $69.15$ & $0.00$\\
				\midrule
				EW$^\S$ & $39.29$ & $65.33$ & $0.5493$ & $0.2263$ & $28.15$ & $23.96$ & $22.09$ & $47.50$ & $61.08$ & $\textcolor{purple}{+0.88}$\\
				GLS & $39.78$ & $65.63$ & $0.5318$ & $0.2272$ & $26.13$ & $21.08$ & $26.57$ & $52.83$ & $65.78$ & $\textcolor{purple}{+5.15}$\\
				RLW$^\S$ & $37.17$ & $63.77$ & $0.5759$ & $0.2410$ & $28.27$ & $24.18$ & $22.26$ & $47.05$ & $60.62$ & $\textcolor{teal}{-2.16}$\\
				\midrule
				UW$^\S$ & $36.87$ & $63.17$ & $0.5446$ & $0.2260$ & $27.04$ & $22.61$ & $23.54$ & $49.05$ & $63.65$  & $\textcolor{purple}{+0.91}$\\
				DWA$^\S$ & $39.11$ & $65.31$ & $0.5510$ & $0.2285$ & $27.61$ & $23.18$ & $24.17$ & $50.18$ & $62.39$  & $\textcolor{purple}{+1.93}$\\
				IMTL-L & $39.78$ & $65.27$ & $0.5408$ & $0.2347$ & $26.26$ & $20.99$ & $26.42$ & $53.03$ & $65.94$ & $\textcolor{purple}{+4.39}$\\
				IGBv2 & $38.03$ & $64.29$ & $0.5489$ & $0.2301$ & $26.94$ & $22.04$ & $24.77$ & $50.91$ & $64.12$ & $\textcolor{purple}{+2.11}$ \\
				\midrule
				MGDA$^\S$ & $30.47$ & $59.90$ & $0.6070$ & $0.2555$ & $\underline{24.88}$ & $19.45$ & $29.18$ & $56.88$ & $\underline{69.36}$  & $\textcolor{teal}{-1.66}$\\
				GradNorm$^*$ & $20.09$ & $52.06$ & $0.7200$ & $0.2800$ & $\textbf{24.83}$ & $\textbf{18.86}$ & $\textbf{30.81}$ & $\textbf{57.94}$ & $\textbf{69.73}$ & $\textcolor{teal}{-11.7}$\\
				PCGrad$^\S$ & $38.06$ & $64.64$ & $0.5550$ & $0.2325$ & $27.41$ & $22.80$ & $23.86$ & $49.83$ & $63.14$  & $\textcolor{purple}{+1.11}$\\
				GradDrop$^\S$ & $39.39$ & $65.12$ & $0.5455$ & $0.2279$ & $27.48$ & $22.96$ & $23.38$ & $49.44$ & $62.87$  & $\textcolor{purple}{+2.07}$\\
				GradVac$^*$ & $37.53$ & $64.35$ & $0.5600$ & $0.2400$ & $27.66$ & $23.38$ & $22.83$ & $48.66$ & $62.21$ & $\textcolor{teal}{-0.49}$\\
				IMTL-G$^\S$ & $39.35$ & $65.60$ & $0.5426$ & $0.2256$ & $26.02$ & $21.19$ & $26.20$ & $53.13$ & $66.24$  & $\textcolor{purple}{+4.77}$\\
				CAGrad$^\sharp$ & $39.18$ & $64.97$ & $0.5379$ & $0.2229$ & $25.42$ & $20.47$ & $27.37$ & $54.73$ & $67.73$  & $\textcolor{purple}{+5.81}$\\
				MTAdam & $39.44$ & $65.73$ & $0.5326$ & $0.2211$ & $27.53$ & $22.70$ & $24.04$ & $49.61$ & $62.69$ & $\textcolor{purple}{+3.21}$\\
				Nash-MTL$^\S$ & $40.13$ & $65.93$ & $\underline{0.5261}$ & $0.2171$ & $25.26$ & $20.08$ & $28.40$ & $55.47$ & $68.15$  & $\textcolor{purple}{+7.65}$\\
				MetaBalance & $39.85$ & $65.13$ & $0.5445$ & $0.2261$ & $27.35$ & $22.66$ & $23.70$ & $49.69$ & $63.09$ & $\textcolor{purple}{+2.67}$\\
				MoCo$^\ddag$ & $40.30$ & $66.07$ & $0.5575$ & $\textbf{0.2135}$ & $26.67$ & $21.83$ & $25.61$ & $51.78$ & $64.85$ & $\textcolor{purple}{+4.85}$\\
				Aligned-MTL$^*$ & $40.82$ & $66.33$ & $0.5300$ & $0.2200$ & $25.19$ & $19.71$ & $28.88$ & $56.23$ & $68.54$ & $\underline{\textcolor{purple}{+8.16}}$\\
				\midrule
				IMTL & $\underline{41.19}$ & $\underline{66.37}$ & $0.5323$ & $0.2237$ & $26.06$ & $20.77$ & $26.76$ & $53.48$ & $66.32$ & $\textcolor{purple}{+6.45}$\\
				DB-MTL (\textbf{ours}) & $\textbf{41.42}$ & $\textbf{66.45}$ & $\textbf{0.5251}$ & $\underline{0.2160}$ & $25.03$ & $19.50$ & $28.72$ & $56.17$ & $68.73$ & $\textcolor{purple}{\textbf{+8.91}}$\\
				\bottomrule
		\end{tabular}}
	\end{table*}
	
	\paragraph{Effectiveness of Gradient-Magnitude Balancing}
	
	Further to the discussion in Section \ref{sec:si-g}, we conduct a comparison between the proposed gradient-magnitude balancing method (i.e., DB-MTL without using logarithm transformation)
	and GradNorm \cite{chen2018gradnorm} on four datasets: \textit{NYUv2}, \textit{Cityscapes}, \textit{Office-31}, and \textit{Office-Home}. As can be seen from Figure \ref{fig:grad_per}, the proposed method consistently achieves better performance than GradNorm in terms of average $\Delta_{\mathrm{p}}$ on all datasets, demonstrating its effectiveness.
	
	\subsection{Sensitivity Analysis} \label{sec:sensitivity_analysis}
	
	\paragraph{Effect of MTL Architecture}
	The proposed DB-MTL is agnostic to the choice of MTL architectures. In this section,  we demonstrate this by evaluating DB-MTL on \textit{NYUv2} using two more MTL architectures:  \textit{Cross-stitch} \cite{MisraSGH16} and  \textit{MTAN} \cite{ljd19}. We compare with  GLS \cite{chennupati2019multinet} and IGBv2 \cite{dai2023improvable}, which perform well in Table \ref{tbl:mtl-nyu_dmtl}. The implementation details are the same as in Section \ref{sec:nyu}. 
	
	Figure \ref{fig:arch} shows each task's improvement performance $\Delta_{\mathrm{p},t}$. For \textit{Cross-stitch} (Figure \ref{fig:arch_cross}), DB-MTL performs the best on all tasks.
	For \textit{MTAN} (Figure \ref{fig:arch_mtan}), all the MTL methods (GLS, IGBv2, and DB-MTL) perform better than STL on both semantic segmentation and depth estimation, but only DB-MTL achieves comparable performance as STL on the surface normal prediction task.
	
	\paragraph{Effect of Backbone Network} 
	We perform an experiment to evaluate DB-MTL on \textit{NYUv2} with the \textit{SegNet} network \cite{badrinarayanan2017segnet} as the backbone. The implementation details are the same as in Section \ref{sec:nyu}, except that the batch size is set to $2$ and data augmentation is used (following CAGrad \cite{liu2021conflict}). As can be seen from Table \ref{tbl:mtl-nyu_segnet}, DB-MTL again achieves the best performance in terms of average $\Delta_{\mathrm{p}}$.
	
\begin{figure*}[!t]
		\centering
		\includegraphics[width=\linewidth]{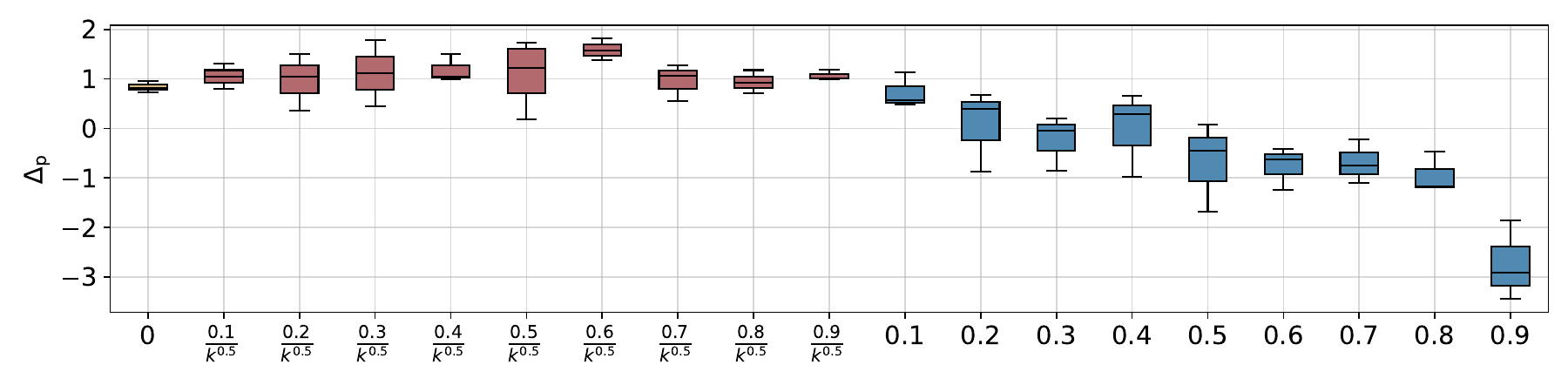} 
		\caption{Effect of EMA's Forgetting Rate $\beta$ in Eq. \eqref{eq:beta} on the \textit{Office-31} dataset. $k$ denotes the number of iterations.}
		% \vskip -0.14in
		\label{fig:31_beta}
	\end{figure*}

	\paragraph{Effect of EMA's Forgetting Rate $\beta$ in Eq. \eqref{eq:beta}}
	As mentioned in Section \ref{sec:nyu}, we perform grid search for $\beta$ over $\{0.1, 0.5, 0.9, \frac{0.1}{k^{0.5}}, \frac{0.5}{k^{0.5}}, \frac{0.9}{k^{0.5}}\}$, where $k$ is the number of iterations. In this experiment, we run DB-MTL on \textit{Office-31} with $\beta\in \{0, 0.1, 0.2, \dots, 0.9, \frac{0.1}{k^{0.5}}, \frac{0.2}{k^{0.5}}, \dots, \frac{0.9}{k^{0.5}}\}$.
	The experimental setup is the same as in Section \ref{sec:image}.
	As can be seen from Figure \ref{fig:31_beta}, the average $\Delta_{\mathrm{p}}$ of DB-MTL is insensitive over a large range of $\beta$ ($\{\frac{0.1}{k^{0.5}}, \frac{0.2}{k^{0.5}}, \dots, \frac{0.9}{k^{0.5}}\}$), and performs better than DB-MTL without EMA ($\beta=0$).
	
	\begin{figure*}[!t]
		\centering
		\includegraphics[width=\linewidth]{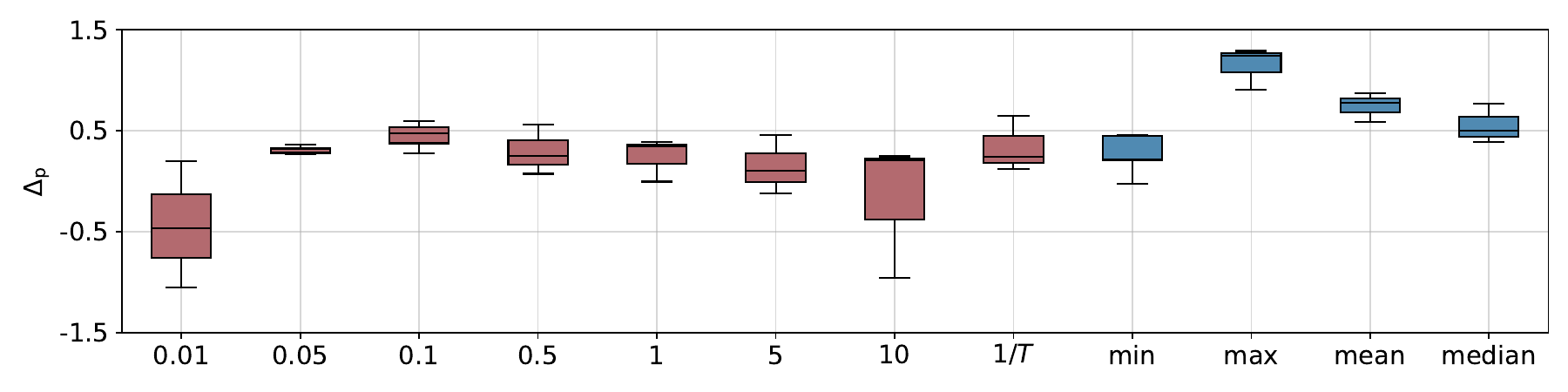} 
		\caption{$\Delta_{\mathrm{p}}$ of different strategies for $\alpha_k$ in Eq. (\ref{eq:si-g}) on the \textit{NYUv2} dataset.
			``min'', ``max'', ``mean'', and ``median'' denote the minimum, maximum, average, and median of $\|\hat{\vg}_{t, k}\|_2~(t=1,\dots,T)$, respectively. $T$ is the number of tasks.}
		% \vskip -0.14in
		\label{fig:alpha_k}
	\end{figure*}

	\begin{figure*}[!t]
		\centering
		% \vskip -.1in
		\includegraphics[width=\linewidth]{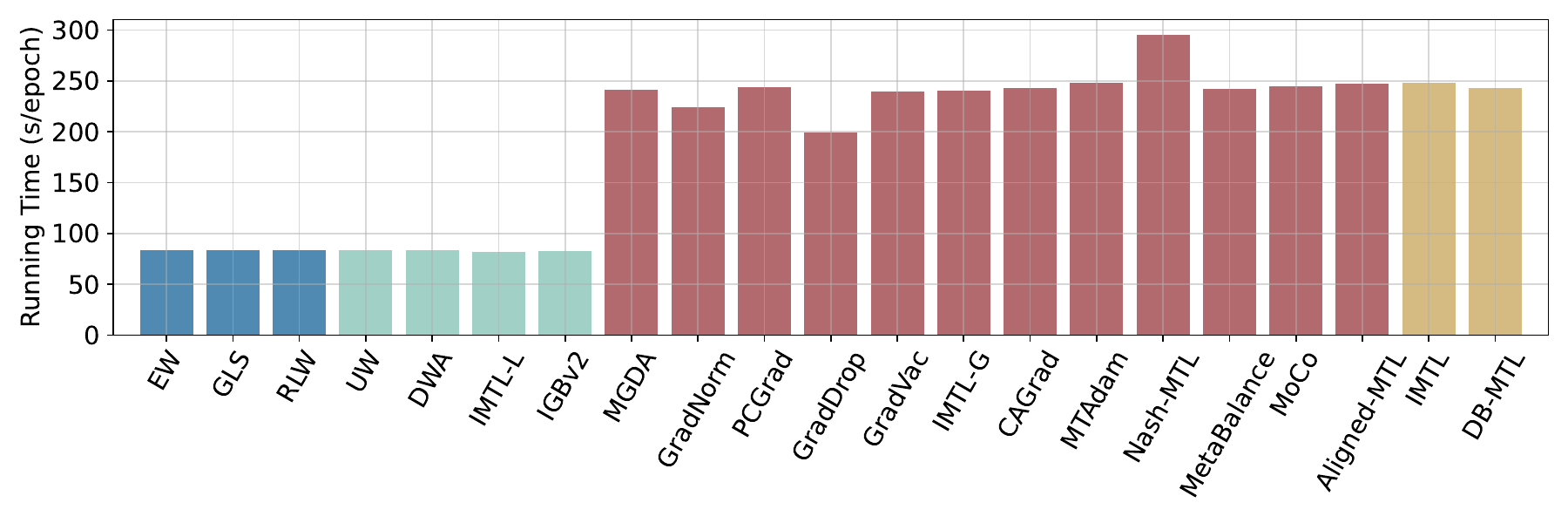} 
		% \vskip -0.15in
		\caption{The running time per epoch averaged 100 repetitions of different methods on \textit{NYUv2} dataset. Cyan, red, yellow, and blue denote loss balancing methods, gradient balancing methods, hybrid balancing methods, and others, respectively.}
		\label{fig:time}
	\end{figure*}
	
	\paragraph{Effect of $\alpha_k$ in Eq. (\ref{eq:si-g})} \label{sec:alpha_k}
	
	In this experiment, we use different settings of $\alpha_k$
	in Eq. (\ref{eq:si-g}), namely, 
	\begin{enumerate*}[(i), series = tobecont, itemjoin = \quad]
		\item constant; 
		\item minimum of $\{\|\hat{\vg}_{t, k}\|_2\}_{t=1}^T$;
		\item maximum of $\{\|\hat{\vg}_{t, k}\|_2\}_{t=1}^T$;
		\item average of $\{\|\hat{\vg}_{t, k}\|_2\}_{t=1}^T$;
		\item median of $\{\|\hat{\vg}_{t, k}\|_2\}_{t=1}^T$. 
	\end{enumerate*}
	Figure \ref{fig:alpha_k} compares the results of these different DB-MTL variants on \textit{NYUv2}. The experimental setup is the same as in Section \ref{sec:nyu}. As can be seen, the maximum-norm strategy performs much better in terms of average
	$\Delta_{\mathrm{p}}$, and thus it is used.
	
	\subsection{Analysis of Training Efficiency} 
	
	Figure \ref{fig:time} shows the per-epoch running time of different MTL methods on \textit{NYUv2} dataset. All methods are run for $100$ epochs on a single NVIDIA GeForce RTX 3090 GPU and the average running time per epoch is reported. As can be seen, DB-MTL has a similar running time as gradient balancing methods (i.e., MGDA \cite{sk18}, GradNorm \cite{chen2018gradnorm}, PCGrad \cite{yu2020gradient}, GradVac \cite{wang2021gradient}, IMTL-G \cite{liu2021imtl}, CAGrad \cite{liu2021conflict}, MTAdam \cite{malkiel2021mtadam}, MetaBalance \cite{he2022metabalance}, MoCo \cite{fernando2023mitigating}, and Aligned-MTL \cite{senushkin2023independent}) and IMTL \cite{liu2021imtl}, but is larger than the loss balancing methods because each task's gradient is computed in every iteration (i.e., step \ref{alg-step:
		task grad} in Algorithm \ref{alg:method}). This is a common disadvantage for gradient balancing methods \cite{wang2021gradient, chen2018gradnorm, navon2022multi, liu2021conflict, sk18,liu2021imtl, yu2020gradient,senushkin2023independent,malkiel2021mtadam,he2022metabalance}. Although DB-MTL is slower than loss balancing methods, it achieves better performance, as shown in Tables \ref{tbl:mtl-nyu_dmtl}, \ref{tbl:mtl-Cityscapes},
	\ref{tbl:qm9}, \ref{tbl:mtl-31-home}, and \ref{tbl:mtl-nyu_segnet}.

	\begin{figure*}[!t]
		\subfloat[\textit{Segmentation task}.]{\label{fig:nyu_train_grad_0}\includegraphics[width=0.3\linewidth]{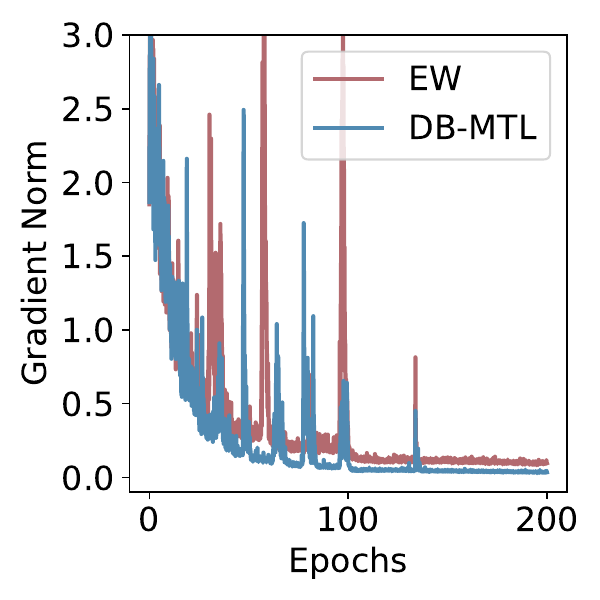}} 
		\hfill
		\subfloat[\textit{Depth estimation task}.]{\label{fig:nyu_train_grad_1}\includegraphics[width=0.3\linewidth]{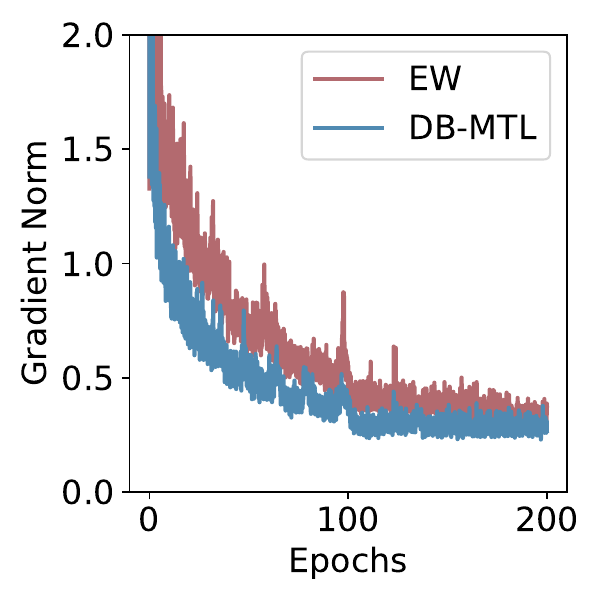}} 
		\hfill
		\subfloat[\textit{Normal prediction task}.]{\label{fig:nyu_train_grad_2}\includegraphics[width=0.3\linewidth]{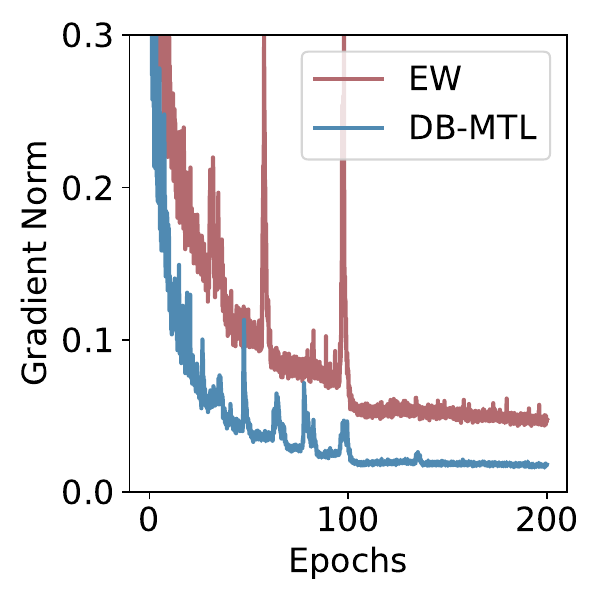}} 
		
		\caption{Gradient norm curves of EW and DB-MTL on the \textit{NYUv2} dataset.}
		\label{fig:train_grad}
	\end{figure*}
	
	\begin{figure*}[!t]
		\subfloat[\textit{Segmentation task}.]{\label{fig:nyu_train_loss_0}\includegraphics[width=0.3\linewidth]{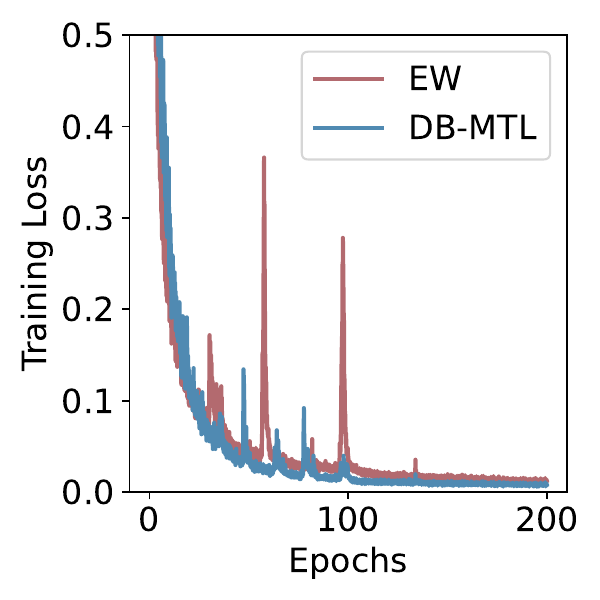}} 
		\hfill
		\subfloat[\textit{Depth estimation task}.]{\label{fig:nyu_train_loss_1}\includegraphics[width=0.3\linewidth]{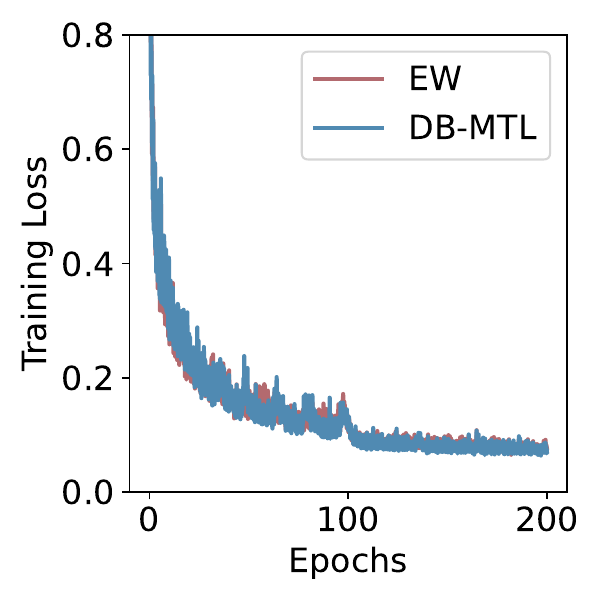}} 
		\hfill
		\subfloat[\textit{Normal prediction task}.]{\label{fig:nyu_train_loss_2}\includegraphics[width=0.3\linewidth]{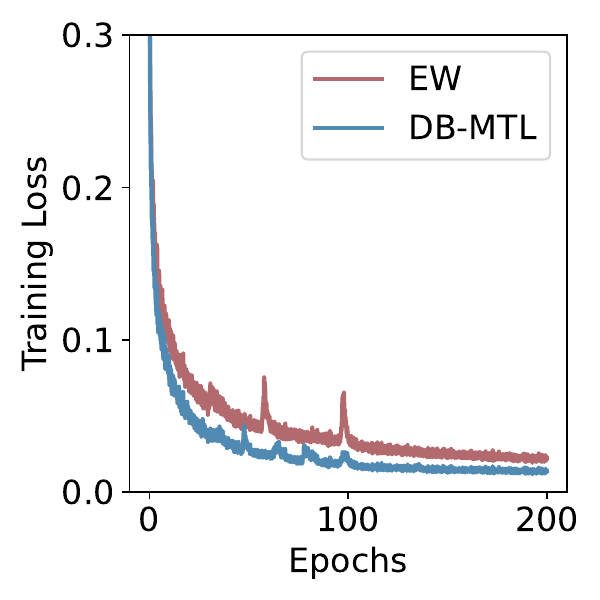}} 
		\caption{Training loss curves of EW and DB-MTL on the \textit{NYUv2} dataset.}
		\label{fig:train_loss}
	\end{figure*}

	\subsection{Analysis of Training Stability}
	Figures \ref{fig:train_grad} and \ref{fig:train_loss} compare the gradient norms $\|\nabla_{\vtheta_{k}}\ell_t({\mathcal{B}}_{t,k};\vtheta_k, \vpsi_{t,k})\|_2$ and training losses of EW and DB-MTL on the \textit{NYUv2} dataset. As can be seen, for each task, the training loss of DB-MTL decreases smoothly and finally converges, and the gradient norm of DB-MTL is much more lower than EW. This indicates the logarithm transformation and maximum-norm strategy do not affect training stability. 
	
	\begin{figure*}[!t]
		\subfloat[\textit{Amazon vs. DSLR}.]{\label{fig:office_grad_cos_0}\includegraphics[width=0.3\linewidth]{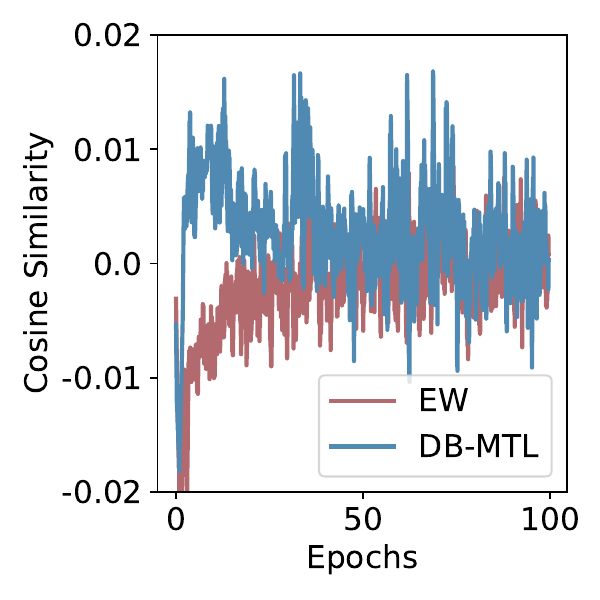}} 
		\hfill
		\subfloat[\textit{Amazon vs. Webcam}.]{\label{fig:office_grad_cos_1}\includegraphics[width=0.3\linewidth]{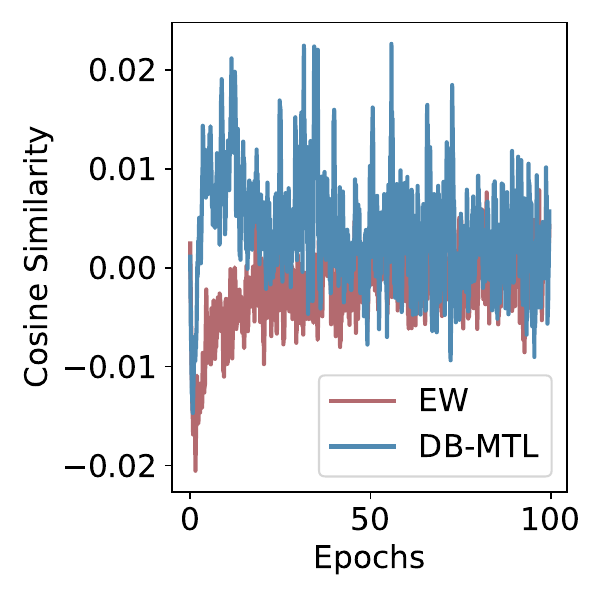}} 
		\hfill
		\subfloat[\textit{DSLR vs. Webcam}.]{\label{fig:office_grad_cos_2}\includegraphics[width=0.3\linewidth]{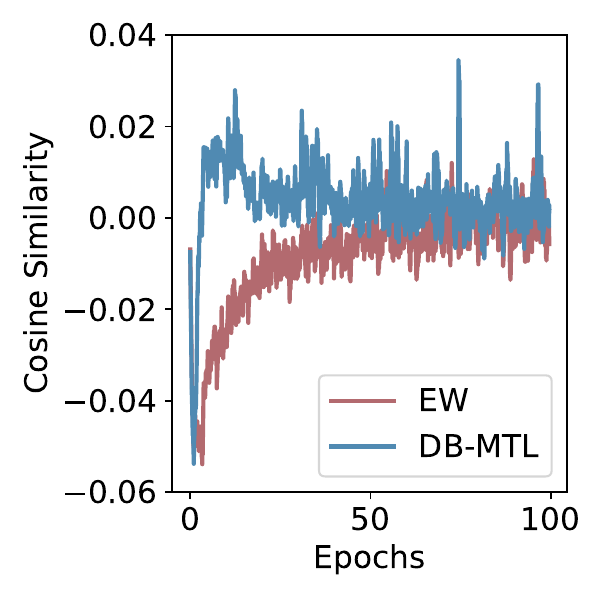}} 
		\caption{Gradient cosine similarity of EW and DB-MTL on the \textit{Office-31} dataset.}
		\label{fig:office_grad_cos}
	\end{figure*}
	
	\subsection{Analysis of Gradient Conflict and Task Imbalance}
	Figures \ref{fig:office_grad_cos} shows the gradient cosine similarity of EW and DB-MTL on the \textit{Office-31} dataset, measuring the gradient conflict and task imbalance \cite{yu2020gradient}. As can be seen, comapred to EW, the cosine similarity of DB-MTL increases faster and then keeps postive during the training process, indicating that DB-MTL can reduce the gradient conflict and improve the task balance.
	
	\section{Conclusion}
	
	In this paper, we alleviate the task-balancing problem in MTL by presenting Dual-Balancing Multi-Task Learning (DB-MTL), a novel
	approach that performs both loss-scale balancing (which makes all task losses have a similar scale via the logarithm transformation)
	and gradient-magnitude balancing (which rescales task gradients to comparable magnitudes using the maximum gradient norm). Extensive experiments on a number of benchmark datasets demonstrate that DB-MTL outperforms the current state-of-the-art. Moreover, the logarithm transformation can also benefit existing gradient balancing methods. {For future work, we will extend our approach to incorporate gradient variance in addition to magnitudes for more refined task weighting, and develop theoretical analysis to provide convergence guarantees and optimality conditions for our method.}
	
	\section*{Acknowledgments}
	
	This work was supported in part by the National Natural Science Foundation of China under Grant No.92370204.
	
	% refs
	\bibliographystyle{elsarticle-num-names} 
	\bibliography{refs}

\end{document}